%% file: main.tex
\documentclass{article}

\usepackage{amsmath}
\usepackage{amsfonts}
\usepackage{graphicx}
\usepackage{booktabs}
\usepackage{dsfont}
\usepackage{subcaption}
\usepackage{makecell}
\usepackage{svg}
\usepackage[preprint]{corl_2026} % Uncomment for pre-prints (e.g., arxiv); This is like ``final'', but will remove the CORL footnote.
\usepackage{xspace}
\newcommand{\ours}{\texttt{Co-GLANCE}\xspace} %Name of framework
\usepackage{wrapfig}
\usepackage{enumitem}
\usepackage{tcolorbox}
\usepackage{listings}
\usepackage{float}
\usepackage{multirow}
\usepackage{pifont}
\newcommand{\cmark}{\ding{51}}  % ✓
\newcommand{\xmark}{\ding{55}}  % ✗
\newcommand{\circletext}[1]{\raisebox{.5pt}{\textcircled{\raisebox{-.9pt} {#1}}}}

\newtcolorbox{promptbox}[1][]{
    colback=gray!10,
    colframe=gray!50,
    fonttitle=\bfseries\small,
    title=#1,
    left=6pt, right=6pt, top=2pt, bottom=2pt,
}

\title{\texttt{Co-GLANCE}: Uncertainty-Aware Active Perception for Heterogeneous Robot Teaming}

\author{
  Michal P. Podolinsky$^{*}$ \, Neel P. Bhatt$^{*}$ \, Pranay Samineni \, Rohan Siva\\
  \textbf{Christian Ellis} \,\textbf{Ufuk Topcu}\\
  The University of Texas at Austin \, $^*$Equal Contribution\\
  \texttt{\{michal.podolinsky,npbhatt,pranay\_s,rohansiva\}@utexas.edu}\\
  \texttt{chrisitan.ellis@austin.utexas.edu, utopcu@utexas.edu}
}

\begin{document}
\maketitle

%===============================================================================

\begin{abstract}

Perceptual uncertainty is a central challenge for heterogeneous robot teams operating in unstructured outdoor environments, where no single viewpoint affords reliable scene understanding.
Perceptual uncertainty, arising from sources such as occlusions, manifests differently across robot viewpoints depending on scene structure.
Detecting and resolving sources of perceptual uncertainty requires both scene-based contextual reasoning and capability-aware robot allocation.
While vision-language models provide strong semantic priors for both, they are computationally prohibitive for onboard inference and lack calibrated uncertainty quantification. 
We introduce \ours{}, a real-time onboard perception and decision-making system for uncertainty resolution in heterogeneous robot teams.
\ours{} distills the semantic reasoning capabilities of a vision-language model into an end-to-end model for occlusion segmentation and robot allocation, eliminating the need for cloud-based inference.
To quantify perceptual uncertainty, \ours{} combines conformal prediction with selective abstention to provide statistically valid coverage guarantees for segmentation, robot allocation, and detection outputs.
These calibrated uncertainty estimates directly trigger active perception, dispatching the most appropriate robot to acquire informative viewpoints and resolve uncertainty.
Across real-world scenarios, \ours{} outperforms cloud-based vision-language model baselines in occlusion segmentation and robot allocation accuracy by 25\% and 36\%, respectively, while reducing per-frame inference latency $350\times$. We also release an air-ground dataset for future research. Code, videos, and dataset available at: \href{https://co-glance.github.io/}{co-glance.github.io}.

\end{abstract}


\input{01-sections/01-introduction}
\input{01-sections/02-related-works}
\input{01-sections/03-methodology}
\input{01-sections/04-experiments}

\input{01-sections/05-conclusion}
% \clearpage
% \input{01-sections/06-appendix}

%===============================================================================

\bibliography{99-Bibliography/michal_references, 99-Bibliography/other_references}  % .bib

\clearpage
\input{01-sections/06-appendix}

\end{document}

%% file: 01-sections/01-introduction.tex
%===============================================================================
\section{Introduction}

\begin{wrapfigure}{r}{0.45\textwidth}
\centering
\vspace{-20pt}
\includegraphics[width=0.42\textwidth]{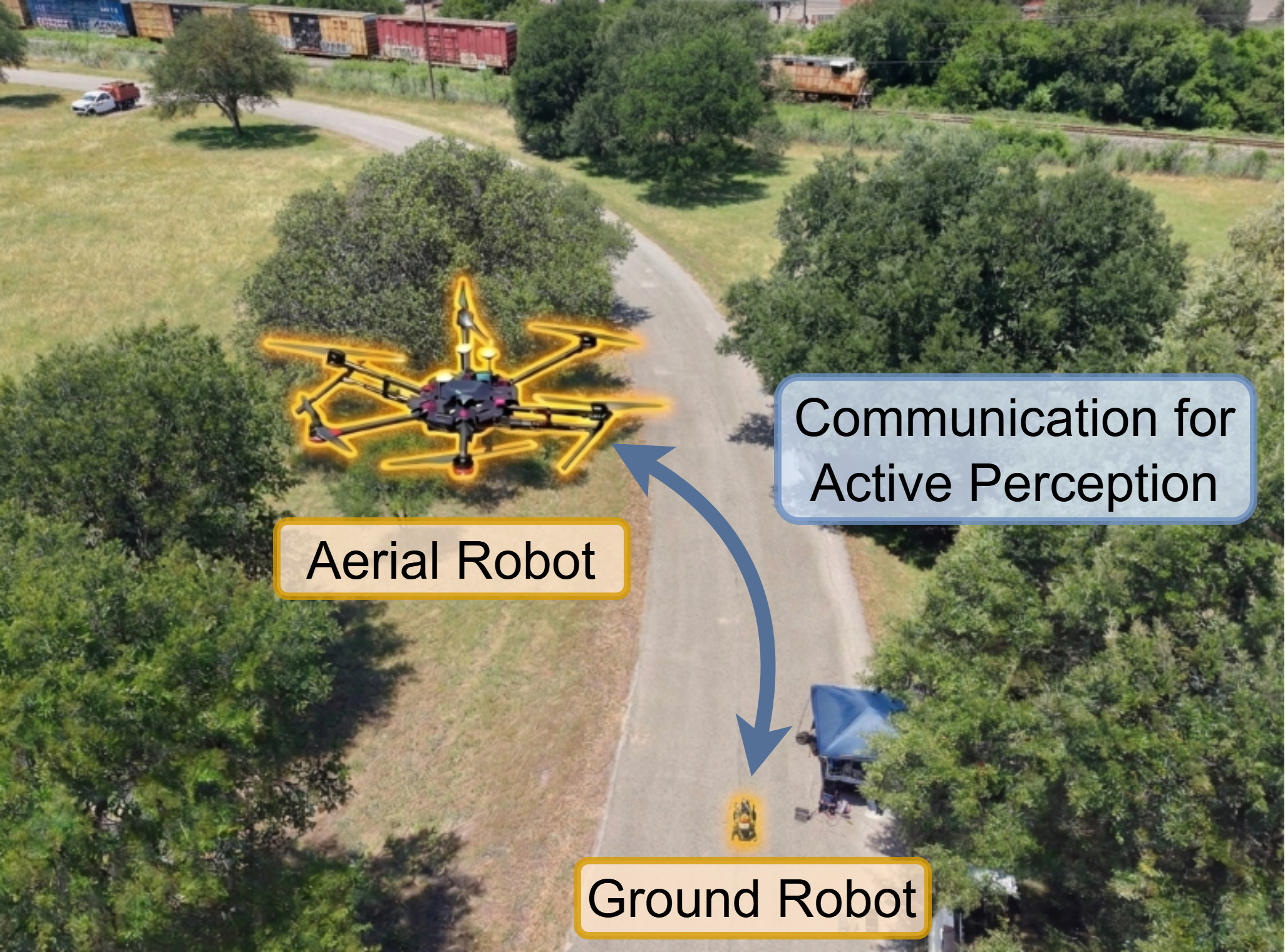}
\vspace{-5pt}
\caption{Air-ground robot teaming setting.}
\label{fig:air-ground-setting}
\vspace{-15pt}  
\end{wrapfigure}

\begin{figure}
    \centering
    \includegraphics[width=\linewidth]{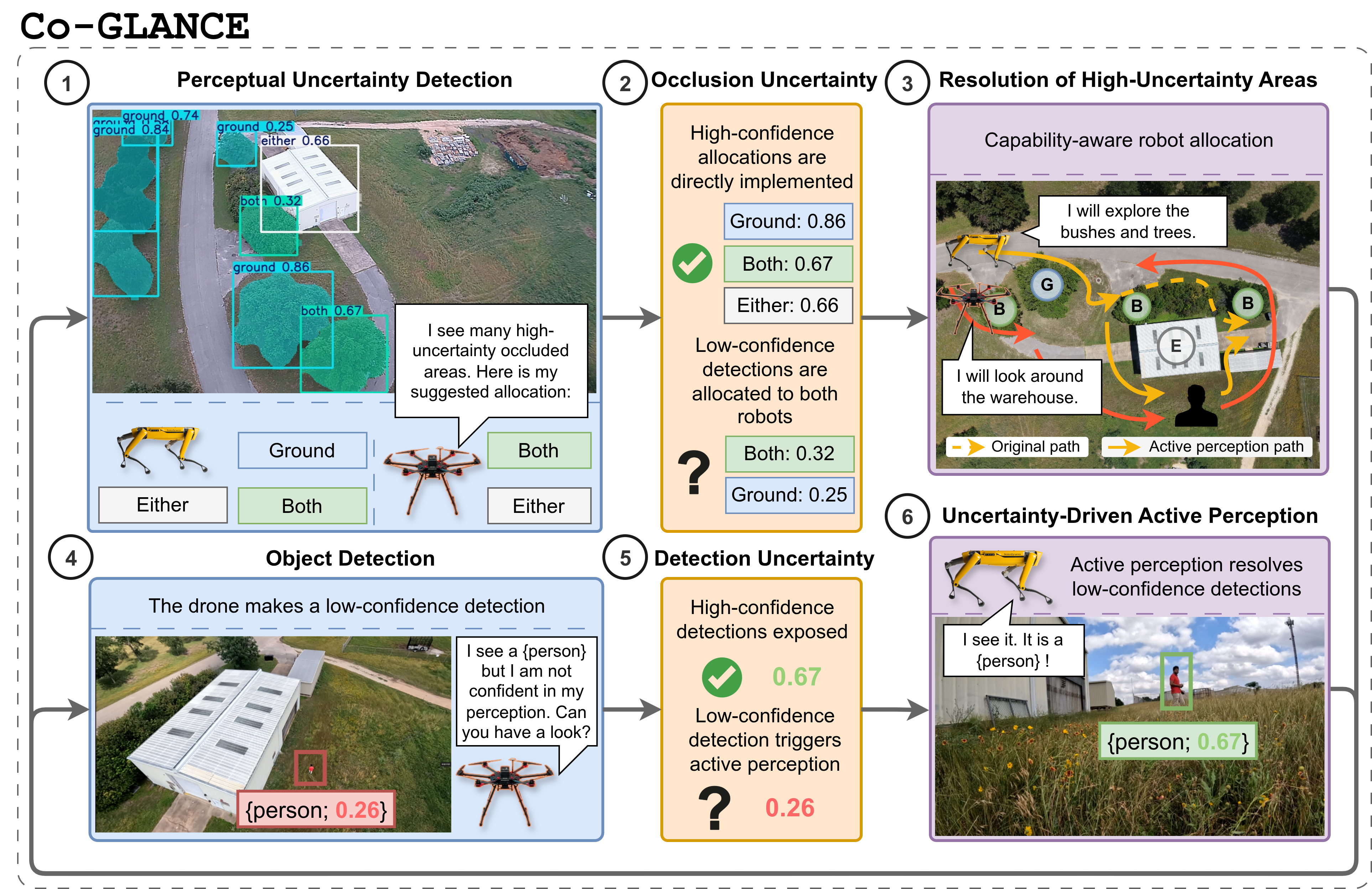}
    \caption{\ours{} \textbf{system overview}: (1) perceptual uncertainty detection, (2) occlusion uncertainty, (3) resolution of high-uncertainty areas, (4) object detection, (5) detection uncertainty, and (6) uncertainty-driven active perception.}
    \label{fig:flowchart}
    \vspace{-10pt}
\end{figure}

Heterogeneous air-ground robot teams provide complementary sensing and mobility capabilities for operating in complex outdoor environments. However, no single viewpoint affords reliable scene understanding in unstructured settings. Perceptual uncertainty arising from occlusions manifests differently across platforms depending on scene geometry and traversal capability: vegetation may obstruct an aerial robot while remaining transparent to a ground robot beneath the canopy, whereas obstacles insignificant from above may fully occlude a ground robot’s view. Hence, detecting and resolving uncertainty requires contextual scene understanding and capability-aware robot coordination.

Recent advances in Vision-Language Models (VLMs) have shown promise for semantic reasoning in heterogeneous robotic systems \cite{ravichandranHeterogeneousRobotCollaboration2025, morilla-cabelloCHORALTraversalAwarePlanning2026, ravichandranSPINEOnlineSemantic2025, claderaAirGroundCollaborationLanguageSpecified2025}. In principle, they can identify ambiguous regions and reason about which platform should resolve them. In practice, however, they are computationally expensive, often require cloud inference, and lack calibrated uncertainty estimates. Existing active perception methods similarly rely on heuristic or uncalibrated confidence, limiting reliability in safety-critical settings.

Conformal prediction provides distribution-free coverage guarantees for uncertainty quantification \cite{angelopoulosGentleIntroductionConformal2022}. However, standard methods produce prediction sets rather than decisions, while selective prediction introduces abstention without resolving uncertainty. In heterogeneous teams, uncertainty must be actionable: deciding when additional sensing is needed and which robot should acquire it.

To address these challenges, we introduce \ours{}, an onboard uncertainty-aware perception and decision-making framework for heterogeneous robot teams. \ours{} distills semantic reasoning from a VLM into a lightweight end-to-end model for occlusion segmentation and robot allocation, removing cloud inference. We further introduce a contextual self-review mechanism that improves consistency of VLM-generated supervision via multi-turn refinement in a cached conversational context. We combine selective abstention with conformal prediction to produce calibrated uncertainty estimates for segmentation, robot allocation, and detection, which directly drive active perception and robot dispatch.

\begin{itemize}[leftmargin=*]
    \item \textbf{Onboard Uncertainty-Aware Perception for Heterogeneous Robot Teams.} \ours{} \ours{} performs real-time occlusion segmentation and robot allocation, improving accuracy by 25\% and 36\% over cloud-based vision-language model baselines.\vspace{-0.2em}

    \item \textbf{Calibrated Uncertainty Estimation for Active Perception.} We combine selective abstention and conformal prediction to provide statistically valid uncertainty guarantees for segmentation, robot allocation, and object detection outputs.\vspace{-0.2em}

    \item \textbf{Contextual Self-Review for VLM Distillation.} We introduce a multi-turn self-review mechanism that improves the consistency of VLM-generated supervision for occlusion reasoning and robot assignment.\vspace{-0.2em}

    \item \textbf{Real-World Deployment and Dataset Release.} We validate \ours{} on aerial-ground robots, achieving $350\times$ lower inference latency and releasing a multimodal air-ground dataset.\vspace{-0.2em}
\end{itemize}

%% file: 01-sections/02-related-works.tex
%===============================================================================

\section{Related Work}
\label{sec:relatedWork}

\textbf{Foundation Models for Multi-Robot Systems.}
Recent work has explored large language and vision-language models for heterogeneous robot coordination and planning \cite{kannanSMARTLLMSmartMultiAgent2024, liuCOHERENTCollaborationHeterogeneous2025, zhuDEXTERLLMDynamicExplainable2025, guptaGeneralizedMissionPlanning2025}. SPINE and SPINE-HT \cite{ravichandranSPINEOnlineSemantic2025, ravichandranHeterogeneousRobotCollaboration2025} extend these ideas to unstructured environments via semantic mapping and feasibility-aware planning. While effective for high-level reasoning and task decomposition, these approaches focus less on uncertainty-aware perception and typically rely on cloud inference without calibrated reliability guarantees. In contrast, our work targets onboard uncertainty-aware perception and capability-aware robot allocation for heterogeneous teams.

\textbf{VLM Distillation for Robotics.}
Recent work uses VLMs as training-time supervisors for lightweight downstream models \cite{xuVLMAD2024,mansourianComprehensiveSurveyKnowledge2025}, transferring multimodal reasoning into compact deployable networks. Applications span autonomous driving, navigation, medical segmentation, and remote perception. Most relevant, \cite{ravichandranDistillingOndeviceLanguage2025} distills language reasoning for onboard inference but still relies on external visual reasoning modules and does not address uncertainty-aware perception. In contrast, we distill both occlusion reasoning and robot allocation into an end-to-end onboard model, while refining pseudo-labels via contextual self-review.

\textbf{Active Perception and Uncertainty Quantification.}
Active perception methods aim to select informative viewpoints to reduce ambiguity \cite{bhattKnowWhereYoure2025, yangHEHAHierarchicalPlanning2025, morilla-cabelloCHORALTraversalAwarePlanning2026}, but often rely on heuristic or uncalibrated confidence signals. Conformal prediction provides distribution-free uncertainty guarantees \cite{vovkAlgorithmicLearningRandom2005, angelopoulosGentleIntroductionConformal2022}, while selective prediction improves reliability via abstention \cite{angelopoulosLearnThenTest2022, feldmanAchievingRiskControl2023}. However, conformal methods yield prediction sets that are difficult to directly use in planning, and selective prediction does not resolve abstentions. We instead combine both within a heterogeneous perception framework where calibrated uncertainty directly drives robot allocation and active perception.

%% file: 01-sections/03-methodology.tex
%===============================================================================

\section{Methodology}
\label{sec:body}
We provide a visual overview of \ours in \autoref{fig:flowchart}. \ours{} combines context-aware occlusion segmentation (\S\ref{sec:vlm_distill}, \autoref{fig:flowchart} \circletext{1}), calibrated perception guarantees (\S\ref{sec:probabilistic_guarantees}, \autoref{fig:flowchart} \circletext{2}\circletext{5}), and capability-aware robot allocation (\S\ref{sec:path_planning}, \autoref{fig:flowchart} \circletext{3}) to resolve visible occlusions under onboard computational constraints. Low-confidence detections (\autoref{fig:flowchart} \circletext{4}) trigger active perception (\autoref{fig:flowchart} \circletext{6}) until a high-confidence observation is confirmed (\autoref{fig:flowchart} \circletext{5}).  

We define: (1) \textit{Occluded area}: A region currently not visible 
because an object lies between it and the active viewpoint. We further require that the occluded area is 
large enough to hypothetically conceal a person in any 
posture.  (2) \textit{Platform allocation label}: Encodes which 
platform is \textit{necessary} to resolve an occlusion, 
not which is currently closest or most convenient. The 
label space is $\{\texttt{ground}, \texttt{both}, 
\texttt{either}\}$, where \texttt{ground} requires 
the ground robot, \texttt{both} requires both robots, and 
\texttt{either} permits flexible assignment. (3) \textit{Active perception}: The deliberate dispatch of an agent to a viewpoint that reduces uncertainty on the identity of an ambiguous object~\citep{bajcsyRevisitingActivePerception2016}.

\subsection{Perceptual Uncertainty Detection}
\label{sec:vlm_distill}

\begin{figure}[h]
    \centering
    \includegraphics[width=\linewidth]{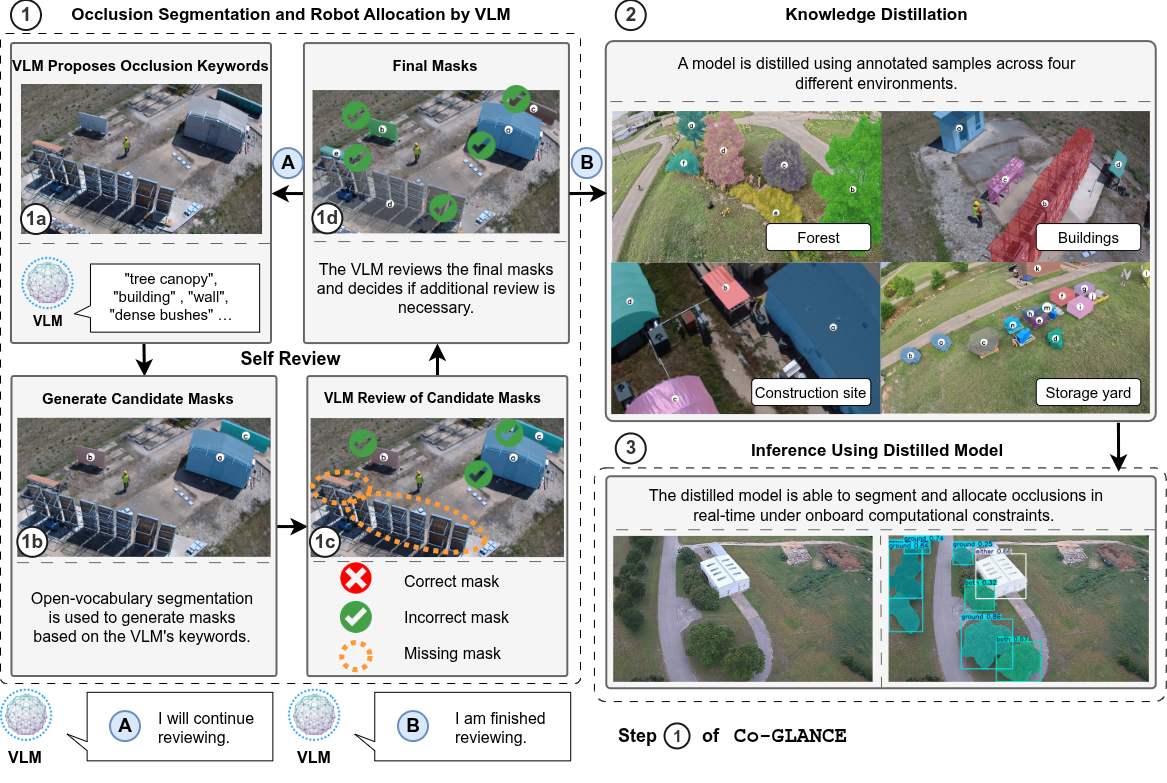}
    \vspace{-20pt}
    \caption{Perceptual uncertainty detection: (1) occlusion segmentation and robot allocation by VLM with self-review, (2) knowledge distillation, and (3) onboard inference using the distilled model.}
    \label{fig:vlm_distillation}
    \vspace{-8pt}
\end{figure}

\ours{} distills occlusion segmentation and platform allocation from a large VLM into a lightweight YOLO-seg-nano model for onboard inference (Figure~\ref{fig:vlm_distillation}). The VLM is prompted to generate occluder keywords from aerial RGB frames (Figure~\ref{fig:vlm_distillation}(1a)), which are passed to an open-vocabulary segmentation model to produce candidate masks (Figure~\ref{fig:vlm_distillation}(1b)). Since the VLM cannot reliably predict how its keywords will be grounded, the resulting masks often misalign with the intended regions or miss occlusions entirely. To address this, a contextual self-review stage presents the candidate masks back to the VLM in a multi-turn conversation, allowing it to remove incorrect masks, refine misaligned regions, propose new keywords, and assign platform labels (Figure~\ref{fig:vlm_distillation}(1c--d)). The distilled model trained on these pseudo-labels segments occlusions and allocates platforms in a single forward pass. Full VLM prompts and additional details are provided in \autoref{app:perception}.

\subsection{Uncertainty Quantification}
\label{sec:probabilistic_guarantees}

\ours{} applies a two-stage uncertainty quantification 
scheme to occlusion segmentation and platform allocation, 
and person detection. The scheme is identical for both prediction types. The risk-controlled stage produces guaranteed 
singletons, where confidence is sufficient, through selective abstention \cite{angelopoulosLearnThenTest2022}; the 
coverage-controlled stage provides calibrated set 
predictions through conformal prediction \cite{angelopoulosGentleIntroductionConformal2022} on the remainder, informing active perception 
while it is underway. The two-stage uncertainty quantification scheme is further detailed in \autoref{app:uq}.

\paragraph{Stage 1 - Risk-Controlled Stage}
\label{par:risk_controlled_stage}

Let $(X_i, Y_i)_{i=1,\ldots,n}$ be i.i.d. sample-label 
pairs, $\hat{Y}(X_i) = \arg\max_y \hat{f}(X_i)$ the 
model prediction, and $\hat{P}(X_i) = \max_y \hat{f}(X_i)$ 
its confidence. For occlusion segmentation and allocation, 
$\hat{Y}(X_i)$ is the correct mask and allocation label 
jointly; for person detection, $\hat{Y}(X_i)$ is the 
correct person mask. The empirical risk is $
    \hat{R}(\lambda) = \frac{1}{n(\lambda)} \sum_{i=1}^{n}
    \mathds{1}\{Y_i \neq \hat{Y}(X_i) \text{ and }
    \hat{P}(X_i) \geq \lambda\}$
where $n(\lambda) = \sum_{i=1}^{n}\mathds{1}
\{\hat{P}(X_i) \geq \lambda\}$. Treating $\hat{R}(\lambda)$ 
as a Binomial random variable, its upper confidence bound 
is $\hat{R}^+(\lambda) = \sup\{r : \mathrm{BinomCDF}
(\hat{R}(\lambda); n(\lambda), r) \geq \delta\}$. We 
select $\hat{\lambda}$ as the last value on a discrete 
grid (fixed sequence testing) where $\hat{R}^+(\lambda) \leq \alpha$, yielding:

\begin{equation}
    \label{eq:selective_accuracy}
    \mathbb{P}\!\big(\mathbb{P}(Y_{\text{true}} = 
    Y_{\text{pred}} \mid \hat{P}(X_{\text{test}}) \geq 
    \hat{\lambda}) \geq 1-\alpha\big) \geq 1-\delta,
\end{equation}

where $\alpha$ is the risk tolerance, $\delta$ the 
confidence parameter, and the outer probability is over 
the calibration set \cite{angelopoulosLearnThenTest2022}. In words: with probability at least 
$1-\delta$, predictions above $\hat{\lambda}$ are correct 
with probability at least $1-\alpha$. Predictions above $\hat{\lambda}$ are used for decision-making; those below $\hat{\lambda}$ are passed 
to stage 2 and generate an active perception request.

\paragraph{Stage 2 - Coverage-Controlled Stage}
\label{par:coverage_controlled_stage}

Let $(X_i, Y_i)_{i=1,\ldots,n}$ be i.i.d. calibration 
samples with $\hat{P}(X_i) < \hat{\lambda}$. We define 
the nonconformity score $S(X_i, Y_i) = 1 - 
\hat{f}_{Y_i}(X_i)$, where higher values indicate worse 
agreement between prediction and label, and compute the 
calibration quantile $\hat{q}$ at level 
$\lceil(n+1)(1-\epsilon)\rceil / n$. The prediction set 
is then:

\begin{equation}
    \hat{C}(X_{\text{test}}) = 
    \left\{y : \hat{f}_y(X_{\text{test}}) \geq 
    1 - \hat{q}\right\},
\end{equation}

which yields the marginal coverage guarantee 
$\mathbb{P}[Y_{\text{test}} \in \hat{C}(X_{\text{test}})] 
\geq 1 - \epsilon$ \cite{angelopoulosGentleIntroductionConformal2022}, 
where probability is taken jointly over calibration and 
test samples. This allows \ours{} to make predictions on all model outputs.

\subsection{Uncertainty Resolution}
\label{sec:path_planning}

\paragraph{Robot Allocation and Routing}
\label{par:routing}

We feed certified allocation labels and agent positions into a Heterogeneous Vehicle Routing Problem (HVRP) \cite{ortools}, minimizing total heuristic travel cost where ground robot traversal is weighted $10\times$ higher than aerial traversal to reflect its slower speed and terrain constraints, with Euclidean distances used as cost approximations. \texttt{Ground}-labeled occlusions are assigned exclusively to the ground robot, \texttt{either}-labeled occlusions are assigned to either robot, and \texttt{both}-labeled or low-confidence occlusions appear in both routes as independent nodes. The ground robot is dispatched to a ground-level viewpoint for each assigned occlusion while the aerial robot performs a fixed circular sweep around each occlusion. See plots in \autoref{app:resolution}.

\paragraph{Active Perception}

\ours{} supports active perception through two channels: uncertain robot allocation and uncertain object detection. For \textit{robot allocation}, stage 1 singleton labels are passed directly to the planner; abstentions trigger a conservative both-agent dispatch, ensuring all high-uncertainty regions are visited. For \textit{object detection}, stage 1 abstentions trigger an active perception request dispatching the ground agent to acquire a closer viewpoint, while stage 2 CP sets inform the system of the most likely object classes present. In our experiments, allocation abstentions are resolved conservatively via both-agent dispatch rather than through CP prediction sets; the sets could enable a dynamic re-routing strategy as agents observe the environment. For object detection, CP sets can used to track possible object identities during active perception. 

%% file: 01-sections/04-experiments.tex
%===============================================================================

\section{Experiments}
\label{sec:result}

We evaluate \ours{} in semi-structured outdoor environments containing both natural vegetation and built infrastructure, where occlusions are highly viewpoint-dependent and often require complementary aerial and ground observations. In addition to validating the system in real-world conditions, we release a multimodal air-ground perception dataset collected in these environments to support future research in heterogeneous robot perception and uncertainty-aware active perception.

\subsection{Experimental Setup}

\paragraph{Robot Platform}
We pair a DJI Matrice 600 aerial robot with a Boston Dynamics Spot quadruped ground robot. The quadruped can traverse rough terrain and dense vegetation that wheeled platforms cannot \cite{biswal2021development}, while the aerial robot provides unconstrained overhead views. Additional information can be found in \autoref{app:setup}.

\paragraph{Environment}
We train the distilled model on over 10000 masks and test it on nearly 200 masks from a site containing the aforementioned vegetation and structural characteristics. In addition, we performed a real-world demonstration of \ours in two scenarios while initializing the robots from opposite sides of the scene, thereby inducing different initial viewing geometries.

\paragraph{Demonstration Scenarios}
In each scenario, \ours{} operates in a single deployment in which both the aerial and ground robots are initialized from the same initial observation and executed simultaneously under identical environmental conditions. The aerial robot is responsible for onboard perception and for generating waypoints for both platforms, while the ground robot executes the received trajectory. We evaluate two initial configurations in the same environment so as to induce different initial viewing geometries without changing the underlying perception problem.

\paragraph{Baselines}
We compare against three baselines for occlusion prediction and robot allocation, all sharing the same routing module for navigation (\S\ref{par:routing}). \textbf{Expert} serves as an empirical upper bound as it has access to ground truth masks. The \textbf{VLM} baseline is a zero-shot, cloud-based system combining ChatGPT-5.4 with Grounding DINO-base \cite{liuGroundingDINOMarrying2023} and SAM 2 Hiera-Large \cite{raviSAM2Segment2024}. \textbf{VLM + Self-Review} augments this pipeline with our contextual self-review mechanism (\S\ref{sec:vlm_distill}).
\paragraph{Metrics}
We evaluate along three axes. \textit{Perception quality:} includes occlusion detection accuracy (fraction of correctly identified occlusions), robot allocation accuracy (agreement with expert assignments), and segmentation performance measured via precision, recall, and F1 score at an IoU threshold of 0.5 between predicted and ground-truth occlusion masks. \textit{Calibration:} characterizes both empirical and guaranteed stage 1 error rates, the abstention rate, and the stage 2 set-valued error rate, quantifying adherence to the target risk while maintaining actionable predictions. \textit{Efficiency:} considers model size, per-frame inference latency, and API token usage.

\subsection{Dataset}
\label{sec:dataset}

\begin{table*}[t]
\centering
\caption{Comparison of datasets for aerial, quadruped, and air-ground collaborative perception.}
\setlength{\tabcolsep}{3.5pt}
\renewcommand{\arraystretch}{1.12}
\resizebox{\textwidth}{!}{
\begin{tabular}{lccccccc}
\toprule
Dataset & Robots & Platforms & Real / Sim & Sensors & Ground Truth & RTK / GPS & Full Bags \\
\midrule
QROD-111~\cite{liObjectTrackingQuadruped2023}
& 1 & Quadruped & Real & RGB & 2D boxes, tracking IDs & \xmark & \xmark \\
EAGLE / CEAR~\cite{zhuEAGLEFirstEvent2024}
& 1 & Quadruped & Real & Event, RGB-D, IMU, LiDAR, Joint Encoder & Robot perception / odometry labels & \xmark & \xmark \\
CDrone~\cite{meierCARLADroneMonocular2024}
& 1 & UAV & Sim & RGB & 2D/3D boxes, tracking, depth, segmentation & \xmark & \xmark \\
UVCPNet / V2U-COO~\cite{wangUVCPNetUAVVehicleCollaborative2024}
& 2 & UAV + vehicle & Sim & Camera & 3D object boxes & \xmark & \xmark \\
M3OT~\cite{nieM3OTMultiDroneMultiModality2025}
& 2 & UAVs & Real & RGB, thermal & 2D boxes, tracking IDs & \xmark & \xmark \\
Griffin~\cite{wangGriffinAerialGroundCooperative2025}
& 2 & UAV + vehicle & Sim & Camera / LiDAR & 3D boxes, tracking, occlusion labels & \xmark & \xmark \\
Our Dataset
& 2 & UAV + quadruped & Real & RGB, RTK GPS, IMU & 2D boxes, tracking IDs& \cmark & \cmark \\
\bottomrule
\end{tabular}}
\label{tab:dataset_comparison}
\vspace{-10pt}
\end{table*}
Real air-ground data is costly to collect, requiring two robots operating outdoors simultaneously with synchronized sensing and metric localization across platforms. As shown in \autoref{tab:dataset_comparison}, many existing datasets sidestep this with simulation, road scenes, or homogeneous UAV teams, and release processed labels rather than raw sensor streams. Our dataset addresses this by providing more than 4000 synchronized aerial and ground frames across several scenarios, recorded with a DJI Matrice 600 and a Boston Dynamics Spot in semi-structured outdoor terrain. Depending on the scenario, available streams include RGB, estimated depth, RTK GPS, IMU; raw ROS~2 bags from both platforms are also released to support evaluation of perception and autonomy stacks beyond static image benchmarks. A full breakdown of dataset scenarios and sensor availability is provided in \autoref{app:dataset}. The dataset is available at \href{https://co-glance.github.io/}{co-glance.github.io}.

\subsection{Quantitative Results}

\paragraph{Distilled Model Performance} We present a comparison of the VLM baseline (with and without self-review) against the distilled model in Table~\ref{tab:model_eval}. The distilled model outperforms the VLM model on precision, recall, F1  score by 22\%, 15\%, and 19\% respectively which highlights that the distilled model can generalize away from noise in the VLM's pseudo-labels. Moreover, self-review improves recall and allocation accuracy by over 15\% by grounding the VLM's keywords against the generated segmented masks. The distilled model trails VLM self-review on allocation accuracy by 5\%, due to the self-review allocation step being harder to distill than the segmentation task itself. However, given the gains on other metrics, \ours provides an overall balanced approach. Additional quantitative results are provided in \autoref{app:results}.

\begin{table}[h]
\centering
\caption{Model-level evaluation on $n=199$ hand-annotated masks across $34$ held-out frames. These frames were not seen during model training.}
\begin{tabular}{lcccc}
\toprule
System & Precision & Recall & F1 & Alloc. Acc.\\
\midrule
VLM (no review)   & 0.458 & 0.543 & 0.497 & 0.694 \\
VLM (self-review) & 0.489 & 0.668 & 0.565 & \textbf{0.850} \\
\ours~(distilled) [ours] & \textbf{0.680} & \textbf{0.693} & \textbf{0.687} & 0.797 \\
\bottomrule
\end{tabular}
\label{tab:model_eval}
\end{table}

\paragraph{Effect of Uncertainty Quantification}

Table~\ref{tab:guarantees} summarizes the effect of incorporating uncertainty quantification across robot allocation and person detection. The guaranteed error rates are chosen to balance error rate against abstention rate, as different tasks may tolerate higher abstention in exchange for lower error. We find that $\alpha = 15\%$ for robot allocation and $\alpha = 20\%$ for person detection offer a reasonable operating point under this tradeoff. The stage 2 error rate $\epsilon$ is chosen to match the stage 1 error rate for increased interoperability. For both tasks, stage 1 ensures that the guaranteed error rate is not exceeded with probability $1 - \delta$ by construction ($\delta = 0.1$). 
 
For robot allocation, stage 1 (cal. $n = 778$ masks, test $n=222$ masks) reduces empirical error relative to the non-UQ baseline while introducing a minimal abstention rate of just above $10\%$. Stage 2 conformal prediction further provides coverage-certified predictions for all abstained instances, ensuring that no inputs are left without a statistically valid decision. A similar trend is observed in person detection (cal. $n=126$ masks, test $n=30$ masks), where selective abstention reduces the effective empirical error rate by 16\% compared to the non-UQ baseline, at the cost of a higher abstention rate due to the increased difficulty of the task and the use of an off-the-shelf detector not trained on aerial viewpoints. Despite this, stage 2 conformal prediction again ensures bounded error on the remaining ambiguous cases, completing the decision pipeline with formal guarantees.

\begin{table}[h]
    \centering
    \caption{Probabilistic guarantees provided by \ours{}. Guarantees hold by construction; empirical error rates are reported for reference, not as validation.}
    \setlength{\tabcolsep}{4pt}
    \begin{tabular}{llcccc}
        \toprule
        & System & \makecell{Error Rate \\ (Stage 1) $\downarrow$} & \makecell{Empirical Rate \\ (Stage 1) $\downarrow$} & \makecell{Abstention \\ Rate} & \makecell{Error Rate \\ (Stage 2) $\downarrow$} \\
        \midrule
        \multirow{2}{*}{Robot allocation}& w/o UQ  & --              & $15\%$ & --    & -- \\
            & w/ UQ   & $\leq 15 \%$ & $12.6\%$ & $10.4\%$ & $\leq 15\%$\\
        \midrule
        \multirow{2}{*}{Person detection}
            & w/o UQ  & --              & $36\%$ & --    & -- \\
            & w/ UQ   & $\leq 20 \%$ & $20 \%$ & $50 \%$ & $\leq 20\%$\\
        \bottomrule
    \end{tabular}
    \label{tab:guarantees}
\end{table}

\subsection{Demonstrations}
\label{sec:demonstrations}

\begin{figure}[t]
    \centering
    \includegraphics[width=\linewidth]{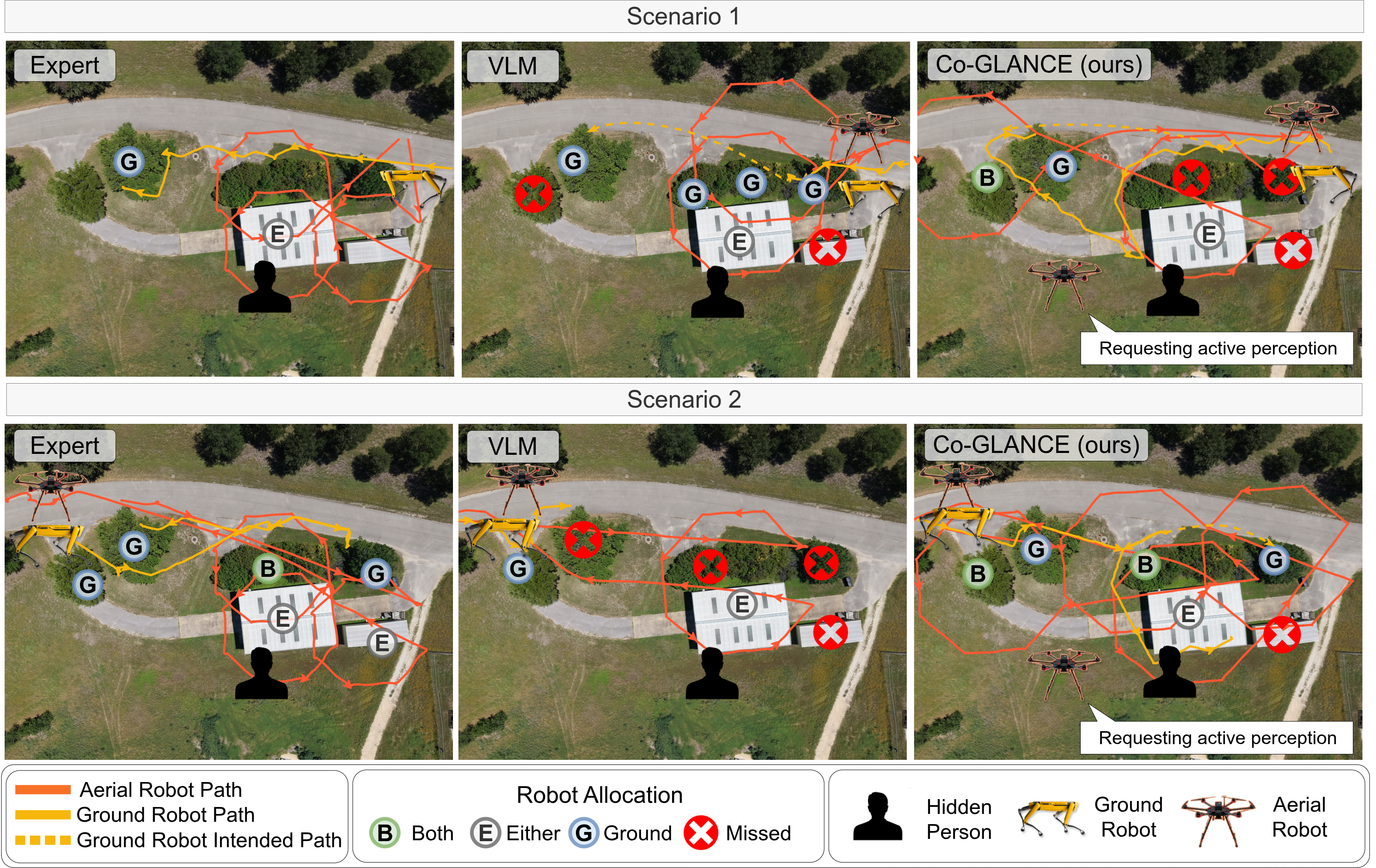}
    \caption{\ours compared with baselines for both scenarios}
    \label{fig:livescenarios}
\end{figure}

\autoref{fig:livescenarios} shows the trajectories of the Expert, VLM, and \ours plans across two real-world air-ground scenarios initialized from opposite sides of the environment. The Expert baseline has access to ground-truth occlusion labels, whereas the VLM baseline relies on cloud-based reasoning without calibrated uncertainty. \ours{} generates onboard occlusion-aware robot allocations and invokes active perception when the aerial robot spots the person with high uncertainty, dispatching the ground robot to acquire complementary viewpoints. Across both scenarios, \ours{} achieves more complete occlusion coverage and invokes active perception to resolve uncertainty while operating entirely onboard. Table~\ref{tab:system_comparison} summarizes end-to-end system performance across two real-world deployment scenarios. \ours{} achieves 25\% higher occlusion detection accuracy and 36\% higher robot allocation accuracy than the cloud-based VLM baseline while operating entirely onboard and without network connectivity. In addition to improving task performance, \ours{} reduces per-frame inference latency by about 350x and eliminates API token usage altogether. These results demonstrate that distilling VLM reasoning into a lightweight onboard model enables practical real-time deployment while preserving strong perception and allocation performance in heterogeneous air-ground settings. Video demonstrations of \ours{} available at: \autoref{app:demos}.

\begin{table}[t]
    \centering
    \caption{System evaluation over two real-world scenarios. Inference time and tokens are per-frame, dataset-wide (\ours{}: 1000 frames, Xavier NX; VLM: $>$1828 frames, RTX 4090).}
    \setlength{\tabcolsep}{4pt}
    \begin{tabular}{lcccc}
        \toprule
        System & Det. Acc. & Alloc. Acc. & \makecell{Infer. Time \\ (ms/fr.)} & \makecell{API Tokens \\ (avg./fr.)} \\
        \midrule
        Expert         & 12/12           & 12/12                & ---               & 0 \\
        VLM (Cloud)    & 6/12            & 5.5/12           & $>$13,000         & 16.7k \\
        % VLM (Onboard)       & Simulated    & 6/12            & ---             & 5.9   & 2.7k & --- & --- \\
        \ours [ours]   & \textbf{7.5/12} & \textbf{7.5/12} & \textbf{37}   & \textbf{0} \\
        \bottomrule
    \end{tabular}
    \vspace{-12px}
    \label{tab:system_comparison}
\end{table}

%% file: 01-sections/05-conclusion.tex
%===============================================================================
\section{Conclusion}

We introduced \ours{}, an uncertainty-aware active perception framework for heterogeneous air-ground robot teams. By distilling semantic reasoning from vision-language models into a lightweight onboard model, \ours{} enables real-time occlusion segmentation, robot allocation, and uncertainty-driven active perception without cloud connectivity. Experimental results demonstrate improved perception and allocation performance over cloud-based VLM baselines while reducing inference latency by approximately $350\times$. We additionally release a multimodal air-ground perception dataset to support future research.

\section{Limitations and Future Work}

While \ours{} demonstrates strong real-world performance, several limitations remain:
\textbf{(1)} the system is evaluated in semi-structured outdoor environments with a limited number of robots,
\textbf{(2)} the active perception policy is heuristic rather than jointly optimized with downstream planning objectives, and
\textbf{(3)} the distilled model may inherit biases from the VLM-generated pseudo-labels used during training.
Future work will explore larger heterogeneous robot teams, planner-aware active perception, and multimodal uncertainty fusion for long-horizon autonomous operation.

%% file: 01-sections/06-appendix.tex
\appendix

\section{Additional Information on Perceptual Uncertainty Detection}
\label{app:perception}

\paragraph{VLM Prompts} The VLM baseline without self-review (\autoref{tab:model_eval}) uses the prompt in \autoref{app:prompt_aerial_single_pass}. The VLM baseline with self-review uses the prompt in \autoref{app:prompt_aerial_initial} to detect occlusions and the prompts in \autoref{app:prompt_aerial_review_1} and in \autoref{app:prompt_aerial_review_2} to review them and allocate the robots.  

\paragraph{Processing Time Breakdown} The self-review pipeline (see \autoref{sec:vlm_distill}) comprises three components: ChatGPT-5.4 as the VLM, Grounding DINO-base \cite{liuGroundingDINOMarrying2023} for open-vocabulary detection, and SAM 2 Hiera-Large \cite{raviSAM2Segment2024} for mask segmentation; hardware specifications are provided in \autoref{tab:hardware}. As shown in \autoref{tab:pipeline_timing}, the VLM dominates processing time, accounting for over 93\% of the total. Open-vocabulary detection and segmentation are invoked an average of 2.39 times per frame, where each call corresponds to one keyword or keyword pair. The VLM also triggers more than two self-review passes per frame on average, reflecting the difficulty of selecting keywords for open-vocabulary segmentation. Together, these observations motivate distilling the VLM's reasoning into a lightweight onboard model, reducing per-frame inference from over 13 seconds to real-time operation.

\begin{table}[h]
\centering
\caption{Timing breakdown of the contextual self-review distillation pipeline ($n=1,828$ frames). $^*$Averaged over 1,828 frames, accounting for multiple detection, segmentation, and review passes per frame.}
\label{tab:pipeline_timing}
\resizebox{\columnwidth}{!}{%
\begin{tabular}{lrrrrrr}
\toprule
\textbf{Stage} & \textbf{Count} & \textbf{Calls/Frame} & \textbf{Avg (s)} & \textbf{Min (s)} & \textbf{Max (s)} & \textbf{Total (s)} \\
\midrule
Pass 1 (VLM Annotation)        & 1{,}828 & 1.00 &  3.961 &  1.638 & 41.306 &  7{,}241.2 \\
Open Vocabulary Detection      & 4{,}365 & 2.39 &  0.264 &  0.130 &  2.912 &  1{,}152.6 \\
Open Vocabulary Segmentation   & 4{,}365 & 2.39 &  0.098 &  0.000 &  0.294 &    429.1 \\
Pass 2 (Self-Review)           & 3{,}770 & 2.06 &  4.107 &  2.031 & 24.127 & 15{,}482.5 \\
\midrule
Total (average$^*$ / frame)               & - & - &  &  - & - & 13.344$^*$ \\
\bottomrule
\end{tabular}%
}
\end{table}

\begin{table}[h]
\centering
\caption{Development machine hardware and software specifications.}
\label{tab:hardware}
\begin{tabular}{ll}
\toprule
\textbf{Component} & \textbf{Specification} \\
\midrule
CPU  & Intel Core i9-13900K (24-core)\\
GPU  & NVIDIA RTX 4090 \\
RAM  & 32 GB \\
OS   & Ubuntu 22.04.5 LTS (Jammy Jellyfish) \\
Kernel & Linux 6.8.0-124-generic x86\_64 \\
\bottomrule
\end{tabular}
\end{table}

\paragraph{Class Distribution} \autoref{tab:class_distribution} shows that the class distribution is skewed toward \{\texttt{either}\} (54.7\%), followed by \{\texttt{ground}\}
(32.2\%) and \{\texttt{both}\} (13.1\%). This distribution is expected: most occlusions can be resolved by either platform repositioning independently, a smaller subset requires ground-level inspection specifically, and the fewest cases demand simultaneous coverage from both robots.

\begin{figure}[H]
\begin{promptbox}[Single-Pass Occlusion Detection and Robot Allocation Prompt]
\begin{lstlisting}[basicstyle=\ttfamily\scriptsize, breaklines=true]
You are analyzing an aerial drone image to identify true physical occlusions: visible objects or structures that block the drone's line of sight to meaningful hidden space behind, under, or inside them.

PLATFORM CONTEXT:
- The drone flies at approximately 10 meters altitude with its camera pointing diagonally downward at roughly 45 degrees from horizontal. Small drone movements of 1-2 meters change the viewpoint very little -- only large repositioning meaningfully changes what is visible.
- A quadruped ground robot operates at ground level with a forward-facing camera at approximately 0.5 meters height. It can navigate uneven terrain and walk around obstacles, but cannot fly, climb walls, or access elevated surfaces.

Focus on the distinction between "object present" and "occlusion present". A visible object is only an occlusion when it hides meaningful space.

VALID OCCLUSION:
- A physical object, vegetation mass, structure, vehicle, or terrain form blocks line of sight.
- The hidden space is large enough to conceal a standing or crouching person. This is a size constraint only -- flag any hidden space large enough to contain a person regardless of whether you expect a person to actually be there. Do not use likelihood of human presence as a criterion.
- The hidden space is hidden by geometry, not by blur, distance, crop, darkness or glare.

NOT AN OCCLUSION:
- Open ground, pavement, grass, dirt, sky, water, shadows, markings, reflections, or image borders.
- Thin poles, wires, signs, sparse branches, isolated trunks, small rocks, low plants, or flat visible surfaces.

FROM AERIAL VIEW, FLAG:
- Tree canopy or dense foliage hiding ground underneath.
- Dense bushes, hedges, shrub clusters, or tall vegetation hiding interior/ground space.
- Building walls/corners, roof overhangs, bridges, underpasses, covered bays.
- Large vehicles, trailers, containers, or equipment hiding space behind/beside/under them.
- Ditches, gullies, berms, ravines, or steep terrain edges hiding bottom or far-side space.

KEYWORDS:
- Use 2-3 word detector-friendly noun phrases.
- Good: "tree canopy", "dense bushes", "building wall", "roof overhang", "cargo trailer", "ditch bank".
- Bad: "shadow", "open ground", "maybe hidden area", "large dark green tree canopy".

PLATFORM LABELS:
For each detected occlusion assign exactly one label:
- ground: underside/interior/beneath-canopy space not visible from above but inspectable by the ground robot.
- either: either drone or ground robot can independently resolve it, such as a simple wall or vehicle with walk-around/fly-over access.
- both: both platforms are needed; use sparingly for compound or dense occlusions neither platform alone resolves such as foliage plus wall/terrain or a thicket hiding interior space from both views.

Return only:
TEXT_PROMPT = "object one . object two ."
EXPLANATIONS = np.array(["Why object one hides meaningful space.", "Why object two hides meaningful space."])
REQUIRED_PLATFORM = np.array(["label one", "label two"])

TEXT_PROMPT, EXPLANATIONS, and REQUIRED_PLATFORM must all have the same length and be in the same order.
Each REQUIRED_PLATFORM entry must be exactly one of: ground, either, both.

If no true occlusions are visible:
TEXT_PROMPT = ""
EXPLANATIONS = np.array([])
REQUIRED_PLATFORM = np.array([])
\end{lstlisting}
\end{promptbox}
\caption{Prompt for single-pass, end-to-end occlusion segmentation and allocation used for the VLM baseline.}
\label{app:prompt_aerial_single_pass}
\end{figure}

\begin{figure}[H]
\begin{promptbox}[Initial Occlusion Detection Prompt]
\begin{lstlisting}[basicstyle=\ttfamily\scriptsize, breaklines=true]
You are analyzing an aerial drone image to identify true physical occlusions: visible objects or structures that block the drone's line of sight to meaningful hidden space behind, under, or inside them.
PLATFORM CONTEXT:
The drone flies at approximately 10 meters altitude with its camera pointing diagonally downward at roughly 45 degrees from horizontal, giving an oblique forward-downward view of the scene. A quadruped ground robot operates at ground level with a forward-facing camera at approximately 0.5 meters height, giving a near-horizontal ground-level perspective.
Focus on the distinction between "object present" and "occlusion present". A visible object is only an occlusion when it hides meaningful space.
VALID OCCLUSION:
A physical object, vegetation mass, structure, vehicle, or terrain form blocks line of sight. The hidden space is large enough to conceal a standing or crouching person. This is a size constraint only -- flag any hidden space large enough to contain a person regardless of whether you expect a person to actually be there. Do not use likelihood of human presence as a criterion. The hidden space is hidden by geometry, not by blur, distance, crop, darkness, glare, or uncertainty.
NOT AN OCCLUSION:
Open ground, pavement, grass, dirt, sky, water, shadows, markings, reflections, or image borders. Thin poles, wires, signs, sparse branches, isolated trunks, small rocks, low plants, or flat visible surfaces. A detector-friendly object that is visible but does not hide meaningful space.
FROM AERIAL VIEW, FLAG:
Tree canopy or dense foliage hiding ground underneath. Dense bushes, hedges, shrub clusters, or tall vegetation hiding interior/ground space. Building walls/corners, roof overhangs, bridges, underpasses, covered bays. Large vehicles, trailers, containers, or equipment hiding space behind/beside/under them. Ditches, gullies, berms, ravines, or steep terrain edges hiding bottom or far-side space.
KEYWORDS:
Use 2-3 word detector-friendly noun phrases. Good: "tree canopy", "dense bushes", "building wall", "roof overhang", "cargo trailer", "ditch bank". Bad: "shadow", "open ground", "maybe hidden area", "large dark green tree canopy".
Return only:
TEXT_PROMPT = "object one . object two ."
EXPLANATIONS = np.array(["Why object one hides meaningful space.", "Why object two hides meaningful space."])
If no true occlusions are visible:
TEXT_PROMPT = ""
EXPLANATIONS = np.array([])
\end{lstlisting}
\end{promptbox}
\caption{Prompt for the initial open-vocabulary segmentation used in the self-review mechanism.}
\label{app:prompt_aerial_initial}
\end{figure}

\begin{table}[h]
\centering
\caption{Class distribution of platform allocation labels across 8{,}190 mask instances in 1{,}828 frames of the VLM-annotated dataset.}
\label{tab:class_distribution}
\begin{tabular}{lrr}
\toprule
\textbf{Class} & \textbf{Count} & \textbf{\%} \\
\midrule
Either  & 4{,}477 & 54.7\% \\
Ground  & 2{,}641 & 32.2\% \\
Both    & 1{,}072 & 13.1\% \\
\midrule
Total   & 8{,}190 & 100.0\% \\
\bottomrule
\end{tabular}
\end{table}

\begin{figure}[H]
\begin{promptbox}[VLM Self-Review Prompt]
\begin{lstlisting}[basicstyle=\ttfamily\scriptsize, breaklines=true]
You are reviewing segmented occlusion masks on an aerial drone image. Colored regions labeled with letters (a, b, c, ...) mark areas previously identified as potential occlusions from the drone's perspective. There are {n_masks} labeled masks in total.

CONTEXT:
- The drone flies at approximately 10 meters altitude with its camera at roughly 45 degrees from horizontal, giving an oblique downward view
- A ground robot navigates the same environment with its camera at approximately 0.5 meters height, giving a near-horizontal perspective
- At 10 meters altitude, small drone movements (1-2 meters) change the viewpoint very little -- only large repositioning meaningfully changes visibility

PLATFORM LABEL RULES:
- "ground": ground robot alone can resolve it (e.g. beneath a low overhang or inside a tunnel the drone cannot descend into)
- "either": drone OR ground robot alone is sufficient -- the drone can fly over AND the ground robot can walk around to observe it independently
- "both": drone AND ground robot are both needed together -- neither alone is sufficient (e.g. simultaneously beneath a canopy the drone cannot penetrate AND behind a wall the ground robot cannot see over)

GOOD MASK CRITERIA:
- Covers a physically meaningful occluding structure (wall, canopy, building, dense vegetation, vehicle)
- The hidden space behind or beneath it is large enough to conceal a person
- The mask boundary reasonably matches the occluding object's visible extent

BAD MASK CRITERIA:
- Covers open ground, sky, shadows, or flat surfaces with no hidden space behind them
- Mask boundary is clearly misaligned with any real structure (floating region, random patch)
- The occluding object is too thin or small to create meaningful hidden space

KEYWORD RULES (for updated and new keywords only):
- 2-3 words maximum
- Simple, generic terms describing object class and basic shape only
- No color, size qualifiers, or complex descriptions
- Good: "concrete wall", "dense bushes", "tree canopy", "cargo truck"
- Bad: "large grey wall", "overgrown hedge row", "big dark green canopy"

TASK -- perform the following steps in order:

STEP 1: Review each labeled mask (a, b, c, ...) and for each decide:

  KEEP -- the mask is a valid occlusion:
  - Assign exactly one platform label: ground, either, or both
  - For "either" labels, confirm in the explanation that both the drone can fly over AND the ground robot can walk around independently

  REMOVE (bad mask) -- the mask is misaligned or low quality but the occluder is real:
  - Drop the mask and provide a corrected 2-3 word keyword in UPDATED_KEYWORDS to regenerate it

  REMOVE (wrong) -- the mask is not a valid occlusion at all:
  - Drop it entirely, no keyword needed

STEP 2: After reviewing all masks, check whether any occluding structures were missed entirely -- tall grass, dense bushes, trees, walls, buildings, vehicles, or terrain features that create meaningful hidden space according to the context above.
  - If yes: provide new 2-3 word keywords in NEW_KEYWORDS
  - If no: return an empty array for NEW_KEYWORDS

[...]
\end{lstlisting}
\end{promptbox}
\caption{Prompt used in the contextual self-review loop after the initial open-vocabulary segmentation.}
\label{app:prompt_aerial_review_1}
\end{figure}

\begin{figure}[H]
\begin{promptbox}[VLM Self-Review Prompt (Continued)]
\begin{lstlisting}[basicstyle=\ttfamily\scriptsize, breaklines=true]
[...]

OUTPUT FORMAT:
Return ONLY these five Python variable assignments in exactly this format and in this order:

REVIEW = np.array(["one sentence verdict for mask a", "one sentence verdict for mask b", ...])
KEEP = np.array(["a", "c", ...])
REQUIRED_PLATFORM = np.array(["label a", "label c", ...])
UPDATED_KEYWORDS = np.array(["corrected keyword for dropped bad mask", ...])
NEW_KEYWORDS = np.array(["new keyword one", "new keyword two", ...])

Rules:
- REVIEW must have exactly {n_masks} entries, one per mask in alphabetical order, each stating: keep/remove and a brief one sentence reason
- KEEP and REQUIRED_PLATFORM must be the same length and in alphabetical mask order
- UPDATED_KEYWORDS contains corrected keywords only for masks removed as bad quality, not for masks removed as wrong
- NEW_KEYWORDS contains keywords for occluders missed entirely in the original segmentation, or an empty array if none
- Each REQUIRED_PLATFORM entry must be exactly one of: ground, either, both
- Do not hallucinate masks or objects not visible in the image

Example -- image has 3 masks (a, b, c): mask a is a valid tree canopy, mask b is a misaligned wall mask, mask c is a shadow incorrectly flagged. One additional tall grass area was missed.

REVIEW = np.array(["keep - dense canopy creates meaningful hidden space beneath that the drone cannot see through.", "remove bad - mask boundary is misaligned with the wall, but the wall is a real occluder worth regenerating.", "remove wrong - this region is a shadow with no physical hidden space behind it."])
KEEP = np.array(["a"])
REQUIRED_PLATFORM = np.array(["ground"])
UPDATED_KEYWORDS = np.array(["concrete wall"])
NEW_KEYWORDS = np.array(["tall grass"])

If there are no masks to keep and no new or updated keywords, or if all masks are valid, always output REVIEW with one sentence of reasoning per mask regardless -- do not skip or abbreviate it. Every run must produce a REVIEW array with exactly {n_masks} entries:

REVIEW = np.array(["keep - dense canopy creates meaningful hidden space beneath that neither platform can resolve alone.", "remove wrong - this region is a shadow with no physical hidden space behind it.", ...])
KEEP = np.array([])
REQUIRED_PLATFORM = np.array([])
UPDATED_KEYWORDS = np.array([])
NEW_KEYWORDS = np.array([])
\end{lstlisting}
\end{promptbox}
\caption{Prompt used in the contextual self-review loop after the initial open-vocabulary segmentation (Continued).}
\label{app:prompt_aerial_review_2}
\end{figure}

\newpage
\section{Additional Information on Uncertainty Quantification}
\label{app:uq}

\begin{figure}[h]
    \centering
    \includegraphics[width=0.7\linewidth]{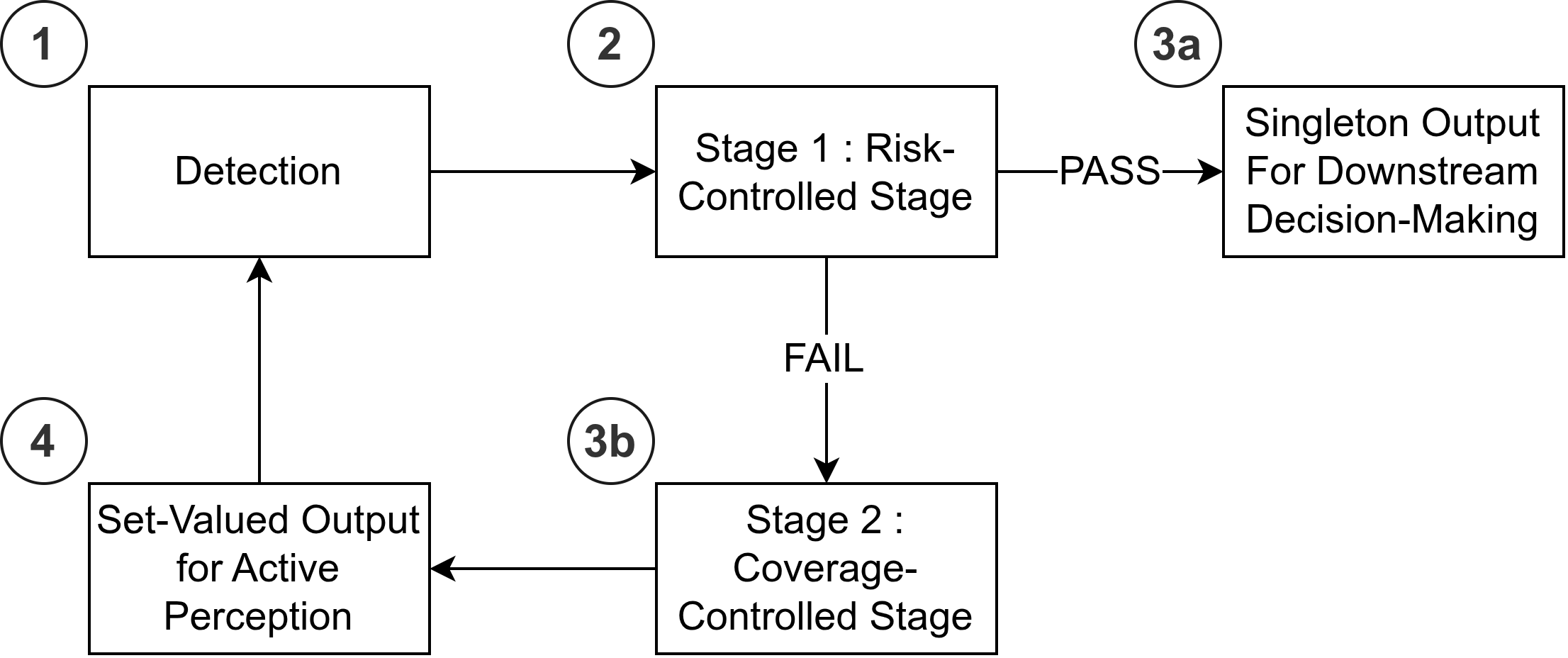}
    \caption{Two-stage uncertainty guarantee scheme. (1) Detection, (2) Risk-controlled stage, (3a) Downstream decision-making informed by singleton output, (3b) Coverage-controlled stage, (4) Active perception informed by set-valued output.}
    \label{fig:guarantee_scheme}
\end{figure}

\paragraph{Two-Stage Guarantee Scheme} To certify model outputs for occlusion segmentation and allocation as well as object detection, \ours{} employs a two-stage guarantee scheme illustrated in \autoref{fig:guarantee_scheme}.

\textit{Example: Object Classification on CIFAR-10 \cite{krizhevskyLearningMultipleLayers}}

To provide an intuitive illustration of the two-stage guarantee scheme, we calibrate both stages on the outputs of a small model trained on CIFAR-10 \cite{krizhevskyLearningMultipleLayers}. \autoref{tab:two-stage-cifar10} reports the calibration thresholds for a scheme calibrated on $n=4000$ samples at the same risk levels as the person detection experiment detailed in \autoref{sec:demonstrations}, along with empirical validation on $n=6000$ samples confirming that the guarantees hold in practice. A detailed example of what this scheme enables is provided below.

\begin{table}[h]
\centering
\caption{Two-stage uncertainty quantification on CIFAR-10. The risk-controlled stage enforces a bounded error rate via selective abstention \cite{angelopoulosLearnThenTest2022}; the coverage-controlled stage enforces set coverage via conformal prediction \cite{angelopoulosGentleIntroductionConformal2022} on the abstained samples.}
\label{tab:two-stage-cifar10}
\setlength{\tabcolsep}{6pt}

\begin{tabular}{@{}lr@{}}
\toprule
\multicolumn{2}{@{}l}{\textbf{Risk-controlled stage}} \\
\midrule
Target error rate $\alpha_\text{sel}$               & 0.200 \\
Failure probability $\delta$                         & 0.100 \\
Threshold $\hat{\lambda}$                            & 0.6403 \\
Empirical error $\hat{R}(\hat{\lambda})$             & 0.1845 \\
Worst-case upper bound $\hat{R}^+(\hat{\lambda})$           & 0.1990 \\
Calibration samples retained (risk-ctrl.)            & 1274 \;(31.9\%) \\
Calibration samples abstained (cov.-ctrl.)           & 2726 \;(68.2\%) \\
\midrule
\multicolumn{2}{@{}l}{\textbf{Coverage-controlled stage}} \\
\midrule
Target miscoverage $\alpha_\text{cp}$                & 0.200 \\
Target coverage $1 - \alpha_\text{cp}$               & 0.800 \\
CP calibration samples                               & 2726 \\
Quantile $\hat{q}$                                   & 0.8855 \\
Softmax inclusion threshold $1 - \hat{q}$            & 0.1145 \\
Achieved test coverage                               & 0.8046 \\
Average prediction set size $|\mathcal{C}|$          & 2.85 \\
\bottomrule
\end{tabular}

\vspace{1em}

\begin{tabular}{@{}lrrrrrl@{}}
\toprule
\multicolumn{7}{@{}l}{\textbf{Empirical validation }} \\
\midrule
Stage & $n$ & Frac. & Accuracy & Err.\ rate & Coverage & \\
\midrule
Risk-controlled  & 1896 & 0.316 & 0.8165 \checkmark & $0.1835$ \checkmark & -- \\
Coverage-controlled & 4104 & 0.684 & -- & -- & $0.8046$ \checkmark \\
\bottomrule
\end{tabular}
\end{table}

\begin{figure}[h]
    \centering
    \begin{subfigure}[b]{\linewidth}
        \centering
        \includegraphics[width=\linewidth]{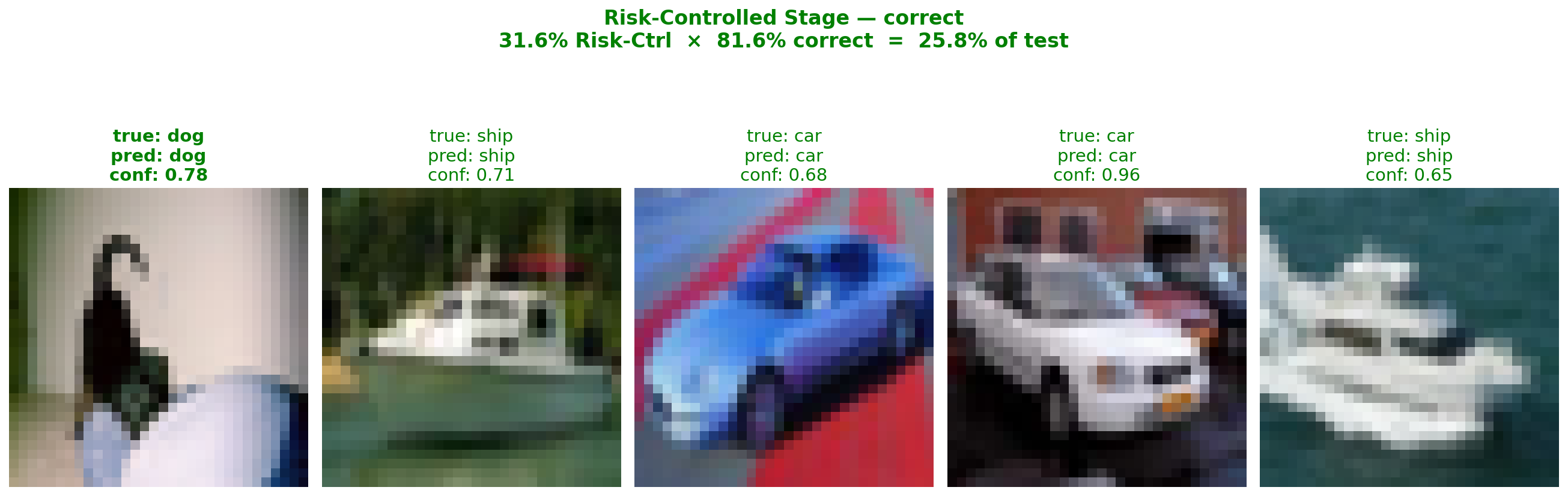}
        \caption{Correct detections at the Risk-Controlled Stage. High-confidence detections are guaranteed to be correct $\geq 80\%$ of the time.}
        \label{fig:ltt-example-correct}
    \end{subfigure}
    \vspace{0.5em}
    \begin{subfigure}[b]{\linewidth}
        \centering
        \includegraphics[width=\linewidth]{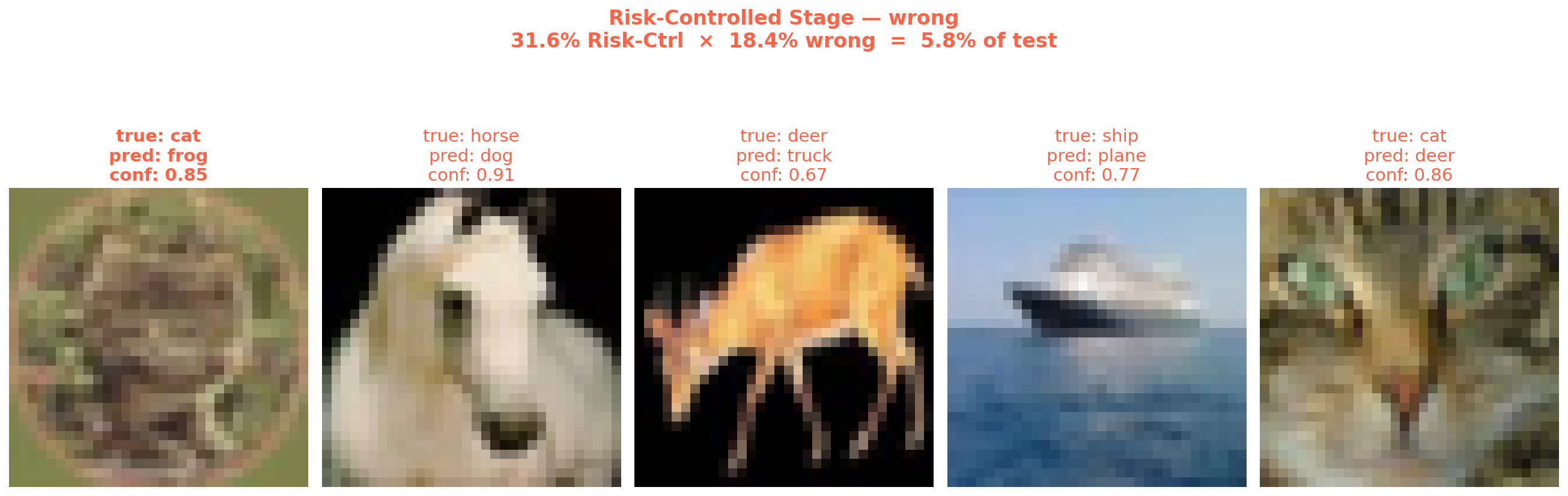}
        \caption{Incorrect detections at the Risk-Controlled Stage.}
        \label{fig:ltt-example-wrong}
    \end{subfigure}
    \vspace{0.5em}
    \begin{subfigure}[b]{\linewidth}
        \centering
        \includegraphics[width=\linewidth]{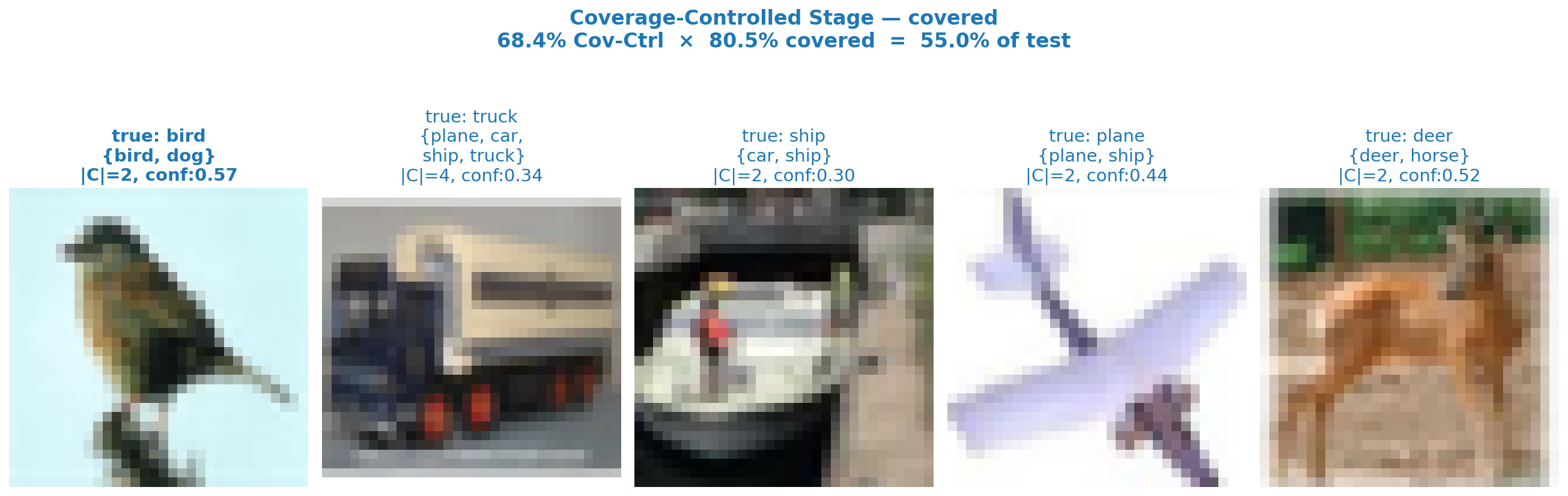}
        \caption{Covered detections at the Coverage-Controlled Stage. Low-confidence detections are guaranteed to be covered $\geq 80\%$ of the time.}
        \label{fig:ltt-example-covered}
    \end{subfigure}
    \vspace{0.5em}
    \begin{subfigure}[b]{\linewidth}
        \centering
        \includegraphics[width=\linewidth]{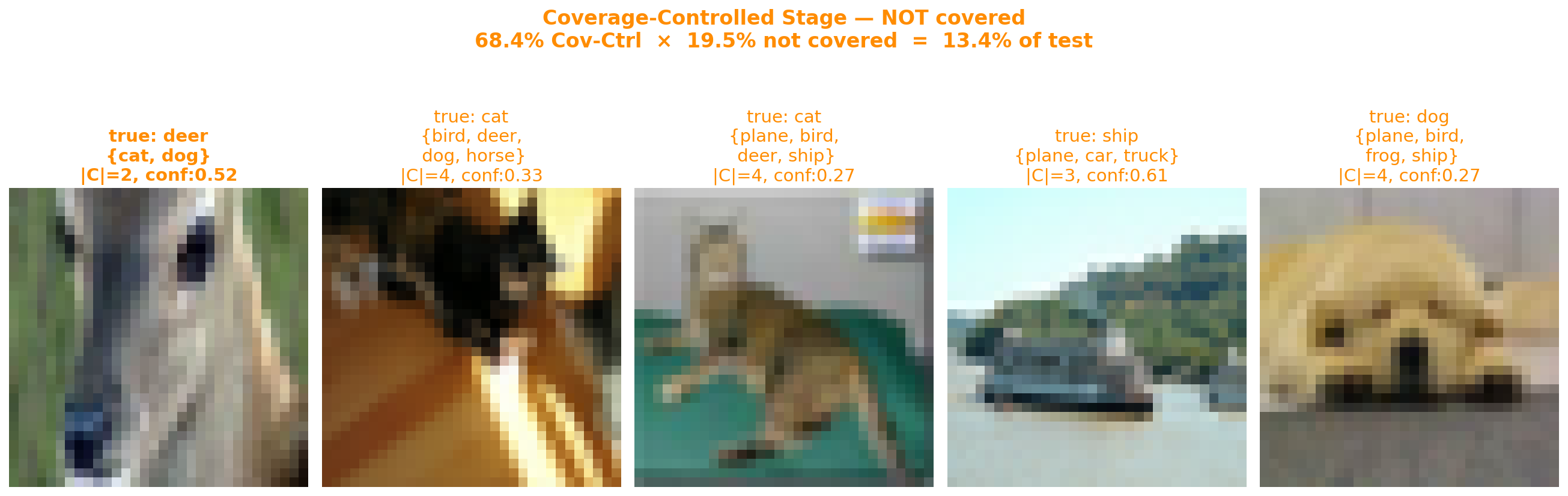}
        \caption{Uncovered detections at the Coverage-Controlled Stage.}
        \label{fig:ltt-example-not-covered}
    \end{subfigure}
    \caption{The two-stage probabilistic guarantee scheme used in \ours{} aims to produce singleton predictions for high-confidence detections and coverage sets for low-confidence detections.}
    \label{fig:ltt-two-tier-examples}
\end{figure}

\begin{itemize}
    \item (1) \textbf{Detection:} While searching for people in a search-and-rescue mission, the aerial robot makes a detection. See examples in \autoref{fig:ltt-example-correct} to \autoref{fig:ltt-example-not-covered}.
    \item (2) \textbf{Risk-controlled stage:} The model prediction (associated with a certain softmax confidence \textit{conf}) is checked against the threshold determined through selective abstention calibration (see \autoref{sec:probabilistic_guarantees}). For this CIFAR-10 example, $\hat{\lambda} = 0.6403$.
    \item (3a) \textit{[conf $\geq$ 0.6403]} The prediction meets the Stage 1 requirement and the detected object is therefore guaranteed to be correctly identified in at least 80\% of cases; active perception is not required. See \autoref{fig:ltt-example-correct}.
    \item (3b) \textit{[conf $<$ 0.6403]} The prediction does not meet the Stage 1 requirement; an active perception request is dispatched to the ground robot. Stage 2 simultaneously produces a calibrated set guaranteed to contain the true label at least 80\% of the time. See \autoref{fig:ltt-example-covered}. This set represents the system's calibrated estimate of the possible semantic class and can be used to guide active perception. For example, if searching for a car, directing active perception toward $\{\texttt{car, ship}\}$ is more informative than toward $\{\texttt{deer, horse}\}$.
    \item (4) \textbf{Active perception:} The ground robot navigates to the requested location and observes the object, reporting back its detection (following the same four-step process until a detection passes the Stage 1 threshold).
\end{itemize}

\textit{Example : Occlusion Segmentation and Allocation}

For occlusion segmentation and allocation, a similar active perception strategy can be employed in which robots do not fully commit to exploring an occlusion until it can be certified with high probability that it is a true occlusion and that it is correctly allocated. In our experiments, we adopt a conservative approach, dispatching both robots to any occlusion that falls below the Stage 1 threshold.

\paragraph{Risk-Controlled Stage Calibration} Following \autoref{sec:probabilistic_guarantees}, we calibrate the Stage 1 threshold on object detections focusing on the ``person'' class and on occlusion segmentation and robot allocation jointly.

\begin{figure}[h]
    \centering
    \includegraphics[width=0.6\linewidth]{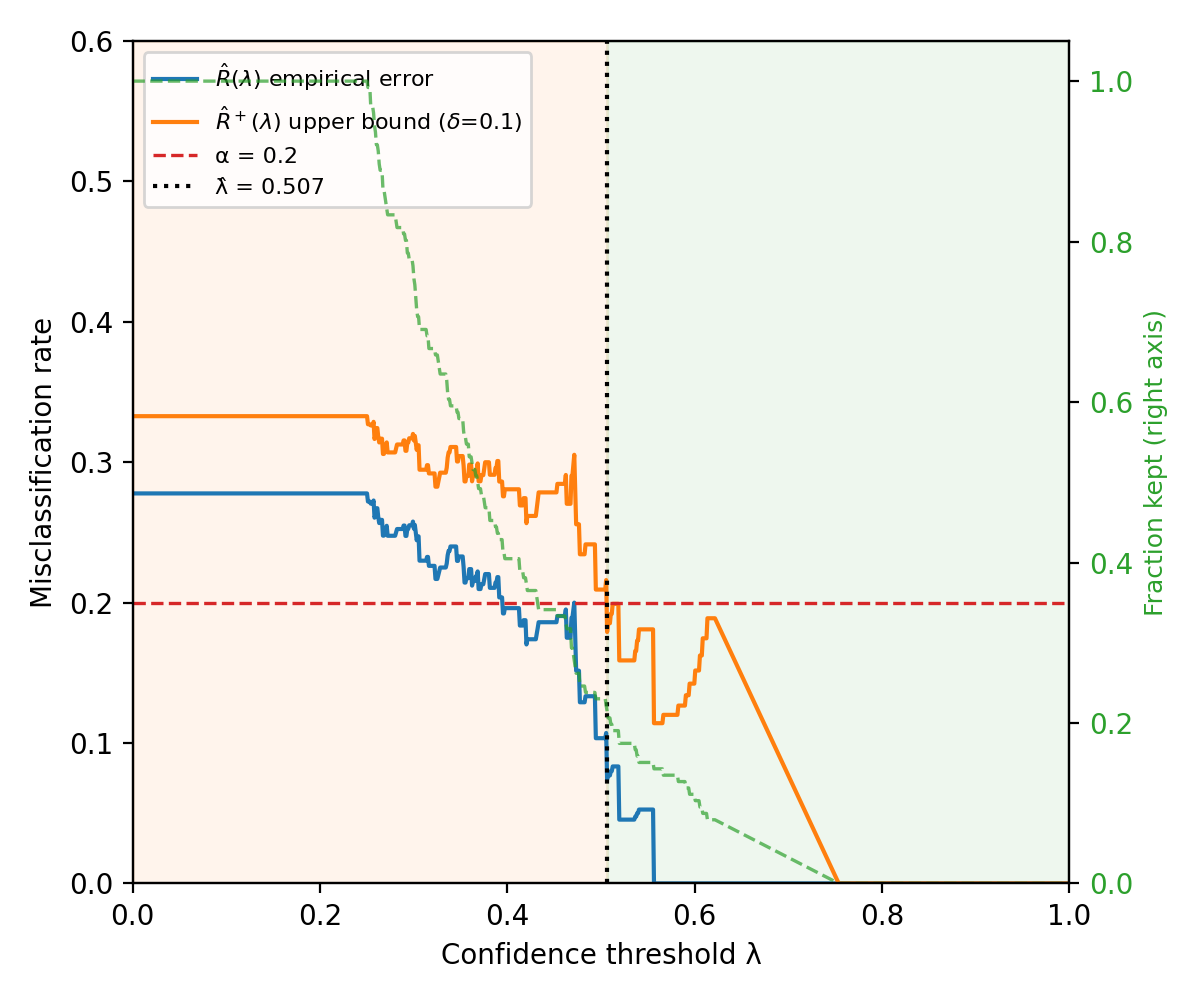}
    \caption{Risk-controlled selective abstention for person detection (aerial viewpoint, YOLO26-nano \cite{glennjocherUltralyticsYOLO262026}).}
    \label{fig:ltt-risk-person}
\end{figure}

\textit{Selective Abstention on Object Detection}

Using a subset of our dataset (see \autoref{sec:dataset}), we calibrate the Stage 1 threshold on person detections using 126 calibration and 30 test detections. Empirical results are provided for reference. The calibration process is illustrated in \autoref{fig:ltt-risk-person}: the blue line shows the empirical error rate, the orange line the worst-case upper bound under the selected $\delta$, the green dashed line the fraction of samples retained at each threshold, and the red dashed line the target error rate $\alpha$. See \cite{angelopoulosGentleIntroductionConformal2022, angelopoulosLearnThenTest2022} for more information on how to read the plots.

\begin{table}[h]
\centering
\caption{Risk-controlled stage calibration results for aerial person detection ($\alpha = 0.2$, $\delta = 0.1$).}
\label{tab:ltt-calibration-aerial}
\begin{tabular}{@{}lr@{}}
\toprule
\multicolumn{2}{@{}l}{\textbf{Calibration setup}} \\
\midrule
Calibration fraction                                         & 0.80 \\
Calibration detections                                       & 126 \\
Test detections                                              & 30 \\
\midrule
\multicolumn{2}{@{}l}{\textbf{Selective abstention results}} \\
\midrule
Target error rate $\alpha$                                   & 0.200 \\
Failure probability $\delta$                                 & 0.100 \\
Threshold $\hat{\lambda}$                                    & 0.5065 \\
Empirical error $\hat{R}(\hat{\lambda})$                     & 0.0741 \\
Worst-case upper bound $\hat{R}^{+}(\hat{\lambda})$                 & 0.1791 \\
Guarantee satisfied $\hat{R}^{+}(\hat{\lambda}) \leq \alpha$ & \checkmark \\
\midrule
\multicolumn{2}{@{}l}{\textbf{Empirical Results}} \\
\midrule
Detections retained                                          & 15\;(50.0\%) \\
Detections abstained                                         & 15\;(50.0\%) \\
Selective precision                                          & 0.800 \\
Selective error                                              & 0.200 \\
Mean confidence (retained)                                   & 0.618 \\
\bottomrule
\end{tabular}
\end{table}

\textit{Selective Abstention on Occlusion Segmentation and Allocation}

\begin{figure}[h]
    \centering
    \includegraphics[width=0.6\linewidth]{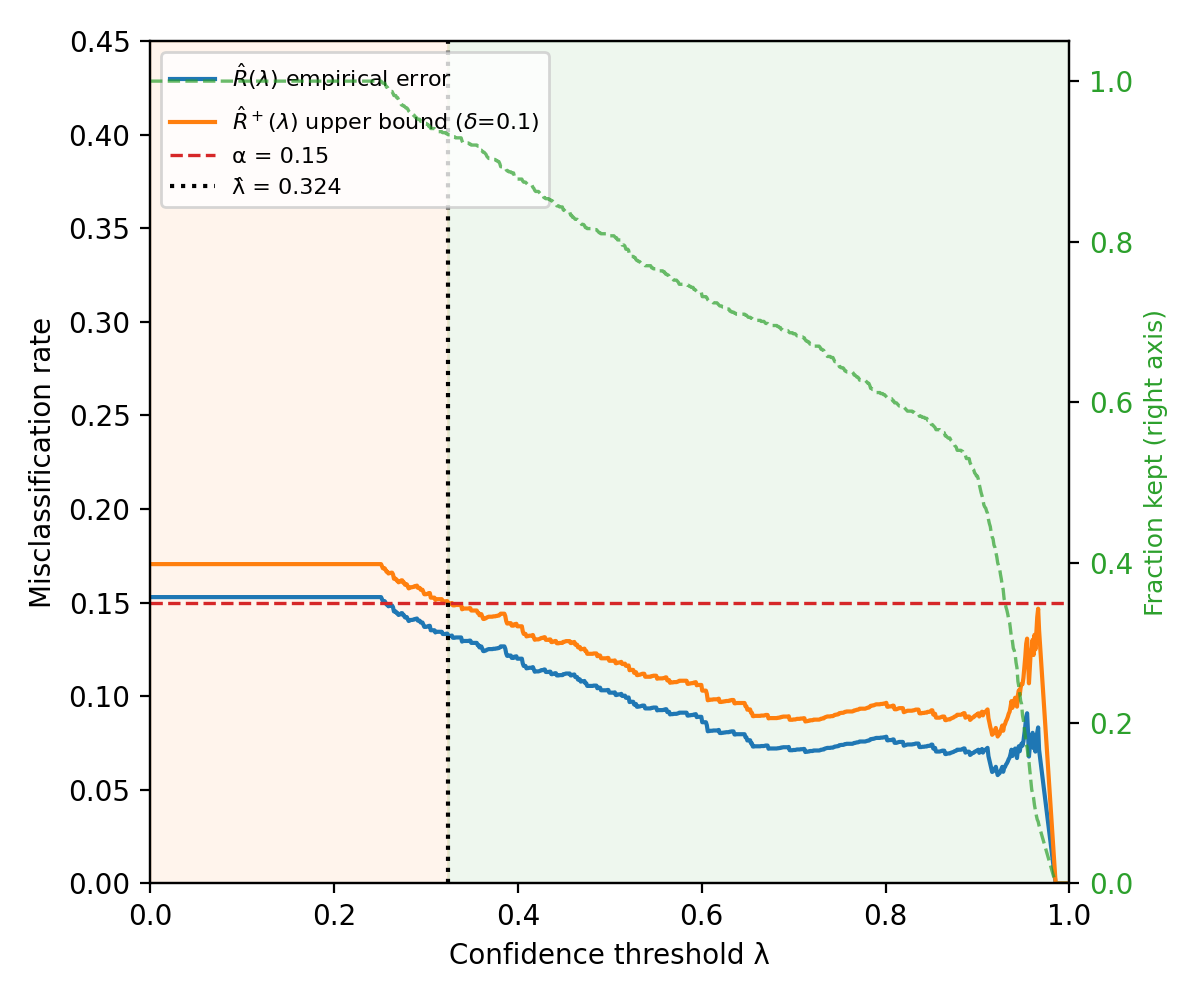}
      \caption{Risk-controlled selective abstention for occlusion segmentation and allocation (\ours{} (distilled)).}
        \label{fig:ltt-risk-occ}
\end{figure}

Using a subset of our dataset (see \autoref{sec:dataset}) and our distilled model (see \autoref{sec:vlm_distill}), we calibrate the Stage 1 threshold on occlusion segmentation and allocation. As noted in \autoref{sec:probabilistic_guarantees}, the risk of incorrect segmentation and incorrect allocation is controlled jointly as a single prediction unit. Calibration parameters and results are reported in \autoref{tab:ltt-calibration-label} and illustrated in \autoref{fig:ltt-risk-occ}.

\begin{table}[h]
\centering
\caption{Risk-controlled stage calibration results for occlusion segmentation and allocation ($\alpha = 0.15$, $\delta = 0.1$).}
\label{tab:ltt-calibration-label}
\begin{tabular}{@{}lr@{}}
\toprule
\multicolumn{2}{@{}l}{\textbf{Calibration setup}} \\
\midrule
Calibration fraction                                         & 0.80 \\
Calibration detections                                       & 778 \\
Test detections                                              & 222 \\
\midrule
\multicolumn{2}{@{}l}{\textbf{Selective abstention results}} \\
\midrule
Target error rate $\alpha$                                   & 0.150 \\
Failure probability $\delta$                                 & 0.100 \\
Threshold $\hat{\lambda}$                                    & 0.3243 \\
Empirical error $\hat{R}(\hat{\lambda})$                     & 0.1322 \\
Worst-case upper bound $\hat{R}^{+}(\hat{\lambda})$                 & 0.1496 \\
Guarantee satisfied $\hat{R}^{+}(\hat{\lambda}) \leq \alpha$ & \checkmark \\
\midrule
\multicolumn{2}{@{}l}{\textbf{Empirical results (aggregate)}} \\
\midrule
Masks retained                                 & 199\;(89.6\%) \\
Masks abstained                                   & 23\;(10.4\%) \\
Selective label accuracy                          & 0.874 \\
Selective label error                            & 0.126 \\
\midrule
\multicolumn{2}{@{}l}{\textit{Per-class breakdown (retained fraction / label accuracy)}} \\
\midrule
\texttt{ground}                                              & 1.000\;/\;0.865 \\
\texttt{both}                                               & 0.947\;/\;0.778 \\
\texttt{either}                                             & 0.871\;/\;0.950 \\
\bottomrule
\end{tabular}
\end{table}

\newpage
\section{Additional Information on Uncertainty Resolution}
\label{app:resolution}
\begin{figure}
    \centering
    \includegraphics[width=\linewidth]{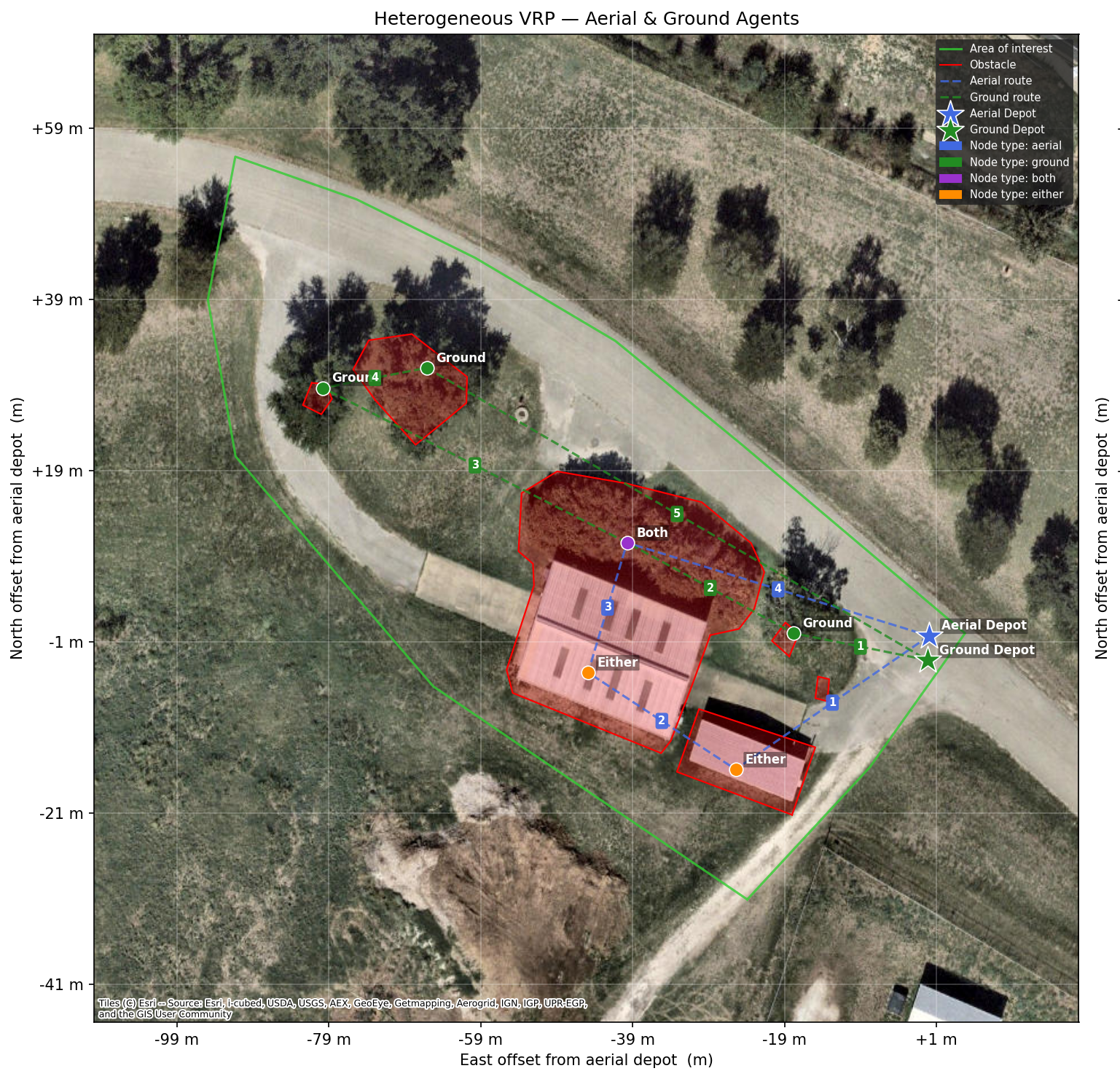}
    \caption{Robot routing, step 1: occlusion allocation (expert demonstration; the expert determines which occlusions to visit, not the robot paths).}
    \label{fig:occlusion-allocation}
\end{figure}

\begin{figure}
    \centering
    \includegraphics[width=\linewidth]{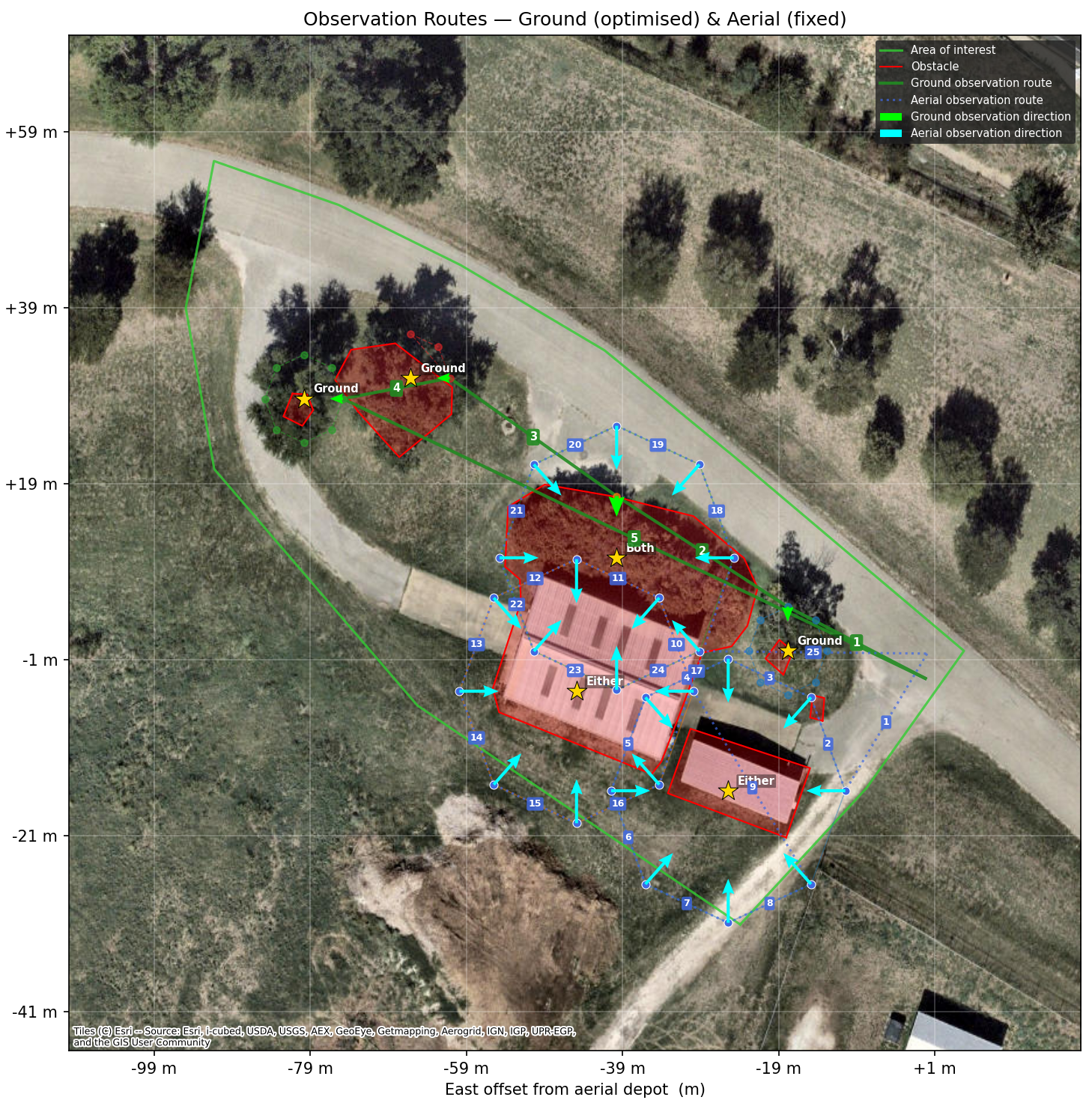}
    \caption{Robot routing, step 2: viewpoint allocation (expert demonstration; the expert determines which occlusions to visit, not the robot paths).}
    \label{fig:viewpoint-allocation}
\end{figure}

\paragraph{Robot Allocation and Routing} Robot allocation and routing is a three-step process, demonstrated visually in \autoref{fig:occlusion-allocation} and \autoref{fig:viewpoint-allocation} on the expert demonstration. Note that the expert is used only to determine the location and type of occlusions, not to specify robot paths. First, each robot is assigned a sequence of occlusions to visit by minimizing a heuristic travel cost using \cite{ortools} (see \autoref{sec:path_planning} and \autoref{fig:occlusion-allocation}). Second, viewpoints are assigned per occlusion: the ground robot receives one viewpoint per occlusion, generated to avoid conflicts with known obstacles identified through satellite imagery; the aerial robot receives eight viewpoints arranged in a circular sweep around each occlusion (see \autoref{fig:viewpoint-allocation}). Third, each robot navigates its assigned sequence of viewpoints. 

\section{Additional Information on Experimental Setup}
\label{app:setup}

\paragraph{Robot Platforms} \ours{} is deployed on two complementary platforms. The ground robot is a Boston Dynamics Spot quadruped, whose legged locomotion enables navigation through dense vegetation and uneven terrain, making it uniquely suited for ground-level occlusion resolution. It runs a NVIDIA Jetson AGX Thor for onboard compute, uses its front-facing RGB cameras for perception, and is equipped with RTK-corrected GPS for localization. The aerial robot is a DJI Matrice 600 Pro, which provides unconstrained overhead coverage of the scene but cannot resolve occlusions beneath vegetation or behind structures. It runs a NVIDIA Jetson Xavier NX, uses an Arducam HQ IMX477 camera for perception, and relies on triple-redundant GPS for localization. Both platforms communicate over local WiFi.

\newpage
\section{Additional Information on the Dataset}
\label{app:dataset}

\ours{} dataset provides over 4,000 synchronized RGB frames (over 2,000 frame pairs) from aerial and ground viewpoints, collected across two outdoor scenarios on semi-structured terrain. Raw ROS 2 bags from both platforms are also released to support evaluation beyond static image benchmarks.

\paragraph{Construction Scenario} A construction worker traverses a construction site tracked by both robots. Ground truth bounding boxes are generated from a GoPro Hero 10 mounted on the ground robot; aerial frames are captured by an Arducam HQ IMX477. The scenario comprises 4 runs and 1,209 annotated frame pairs (\autoref{tab:dataset_runs}): (1) a construction worker standing then walking unoccluded; (2) the worker traversing the construction site in one direction; (3) the same traverse in the opposite direction; (4) a traverse of an adjacent site. The aerial robot loiters overhead across all runs.

\paragraph{Camouflage Scenario} Two camouflage-wearing individuals move through a visually occluded area. Ground truth bounding boxes are generated from a stitched view of the Spot's onboard cameras; aerial frames are captured by a GoPro Hero 13. The scenario comprises 3 runs and 862 annotated frame pairs (\autoref{tab:dataset_runs}): (1) the ground robot follows the individuals through thick brush; (2) the ground robot observes the individuals moving around the brush from an adjacent grassy area; (3) the ground robot follows the individuals while they are occluded by large crates. The aerial robot loiters overhead across all runs.

\begin{table}[hbt!]
    \centering
    \begin{tabular}{llr}
    \toprule
    \textbf{Scenario} & \textbf{Run} & \textbf{Frame Pairs} \\
    \midrule
    Construction & 1 & 118  \\
                 & 2 & 326  \\
                 & 3 & 280  \\
                 & 4 & 485  \\
                 & Total & 1{,}209 \\
    \midrule
    Camouflage   & 1 & 186 \\
                 & 2 & 545 \\
                 & 3 & 131 \\
                 & Total & 862 \\
    \midrule
    Overall Total & & 2{,}071 \\
    \bottomrule
    \end{tabular}
    \caption{Frame pair counts per scenario and run.}
    \label{tab:dataset_runs}
\end{table}

\section{Additional Information on Quantitative Results}
\label{app:results}

To complement \autoref{tab:model_eval}, we provide a per-class breakdown in \autoref{tab:model_eval_perclass}. \ours{} outperforms both baselines on the $\{\texttt{both}\}$ and $\{\texttt{either}\}$ classes across all segmentation metrics. For the $\{\texttt{ground}\}$ class, VLM (self-review) achieves higher precision and F1; however, \ours{} still outperforms VLM (no review) in these cases, confirming that VLM (self-review) represents a considerably stronger baseline. The one exception is allocation accuracy for the $\{\texttt{both}\}$ class, where \ours{} trails both baselines, which we hypothesize reflects the difficulty of distilling the compound scene-level reasoning required to identify occlusions that neither platform can resolve alone.

\begin{table}[h]
\centering
\caption{Per-class model-level evaluation on hand-annotated masks across $34$ held-out frames. These frames were not seen during model training. $^*$\ours{} outperforms the VLM baseline without self-review.}
\begin{tabular}{llcccc}
\toprule
Class & System & Precision & Recall & F1 & Alloc. Acc. \\
\midrule
\multirow{3}{*}{\textit{both}}
 & VLM (no review)        & 0.583          & 0.488          & 0.532          & \textbf{0.667} \\
 & VLM (self-review)      & 0.500          & 0.674                & 0.574          & 0.586          \\
 & \ours~(distilled)      & \textbf{0.689} & \textbf{0.721 }         & \textbf{0.705} & 0.516          \\
\midrule
\multirow{3}{*}{\textit{either}}
 & VLM (no review)        & 0.406          & 0.532          & 0.460          & 0.897          \\
 & VLM (self-review)      & 0.424          & 0.670                   & 0.520          & \textbf{0.932} \\
 & \ours~(distilled)      & \textbf{0.704} & \textbf{0.697}          & \textbf{0.700} & 0.921$^*$          \\
\midrule
\multirow{3}{*}{\textit{ground}}
 & VLM (no review)        & 0.509          & 0.617          & 0.558          & 0.310          \\
 & VLM (self-review)      & \textbf{0.738} & \textbf{0.660} & \textbf{0.697} & \textbf{0.903} \\
 & \ours~(distilled)      & 0.620$^*$          & \textbf{0.660}          & 0.639$^*$          & 0.774$^*$          \\
\bottomrule
\end{tabular}
\label{tab:model_eval_perclass}
\end{table}

\section{Additional Information on Demonstrations}
\label{app:demos}

For video demonstrations, please visit 
\url{https://co-glance.github.io}.